\documentclass[twoside]{article}

\usepackage[accepted]{aistats2022-arxiv}
\usepackage[american]{babel}

\usepackage{natbib} 

\usepackage{mathtools} 
\usepackage{booktabs} 
\usepackage{tikz} 

\usepackage{natbib}
\usepackage[utf8]{inputenc} 
\usepackage[T1]{fontenc}    
\usepackage{hyperref}       
\usepackage{url}            
\usepackage{booktabs}       
\usepackage{amsfonts}       
\usepackage{nicefrac}       
\usepackage{microtype}      
\usepackage{amsthm}
\usepackage{algorithm}
\usepackage{amsmath}
\usepackage{amsthm}
\usepackage{algpseudocode}
\usepackage{nccmath}
\usepackage{subcaption}
\usepackage{graphicx}
\usepackage{wrapfig}
\usepackage{braket}
\usepackage{bm}
\usepackage{etoc}

\usepackage{maths}

\allowdisplaybreaks
\everypar{\looseness=-1	}
\usepackage{xcolor}
\hypersetup{
	colorlinks,
	linkcolor={red!40!gray},
	citecolor={blue!40!gray},
	urlcolor={blue!70!gray}
}

\begin{document}

%

%

\twocolumn[

\aistatstitle{Sensing Cox Processes via Posterior Sampling and Positive Bases}

\aistatsauthor{ Mojm\'ir Mutn\'y \And Andreas Krause}

\aistatsaddress{ ETH Z\"urich \\ \texttt{mojmir.mutny@inf.ethz.ch} \And  ETH Z\"urich \\ \texttt{krausea@inf.ethz.ch}  } ]

\etocdepthtag.toc{mtchapter}
\etocsettagdepth{mtchapter}{none}
\etocsettagdepth{mtappendix}{none}

\begin{abstract}
\looseness -1 We study adaptive sensing of Cox point processes, a widely used model from spatial statistics. We introduce three tasks: maximization of captured events, search for the maximum of the intensity function and learning level sets of the intensity function. We model the intensity function as a sample from a truncated Gaussian process, represented in a specially constructed positive basis. In this basis, the positivity constraint on the intensity function has a simple form. We show how the \emph{minimal description positive basis} can be adapted to the covariance kernel, to non-stationarity and make connections to common positive bases from prior works. Our adaptive sensing algorithms use Langevin dynamics and are based on posterior sampling (\textsc{Cox-Thompson}) and top-two posterior sampling (\textsc{Top2}) principles. With latter, the difference between samples serves as a surrogate to the uncertainty. We demonstrate the approach using examples from environmental monitoring and crime rate modeling, and compare it to the classical Bayesian experimental design approach.
\end{abstract}
\section{INTRODUCTION}
Poisson point processes are extensively used to model event occurrences \citep{Diggle2013b}. They have been successfully applied in crime rate modeling \citep{Shirota2017}, genetic biodiversity \citep{Diggle2013} as well as environmental modeling \citep{Heikkinen1999}. The \emph{intensity function} of a point process is a function whose integral over a region is proportional to the probability of an event occurring in that region. A particularly widespread example of a Poisson point process is the {\em Cox process}, where the intensity function is itself assumed to be random, often modeled as a sample from a Gaussian process \citep{Cox1955}. 

In this work, we address the problem of {\em adaptive sensing of spatio-temporal Cox processes}. Our goal is to adaptively locate and capture events in particular regions of the space -- \emph{sensing regions} -- in order to infer the intensity function, and consequently achieve certain objectives. Depending on the objective, the choice of sensing regions differs and leads to different algorithms. Examples of objectives include identification of the region with the highest incidence of events, identifying level sets of $\lambda$, and capturing the most events subject to the costs of sensing. Specifically, we assume that the intensity {\em varies smoothly} over the sensed domain, which we incorporate into the prior for $\lambda$ by assuming it is \emph{an unknown sample} from a truncated Gaussian Process with a \emph{known covariance kernel $k$}. 

\begin{figure*}[t]
	\centering
	\begin{subfigure}[t]{0.30\textwidth}
		\centering
		\includegraphics[width=\textwidth]{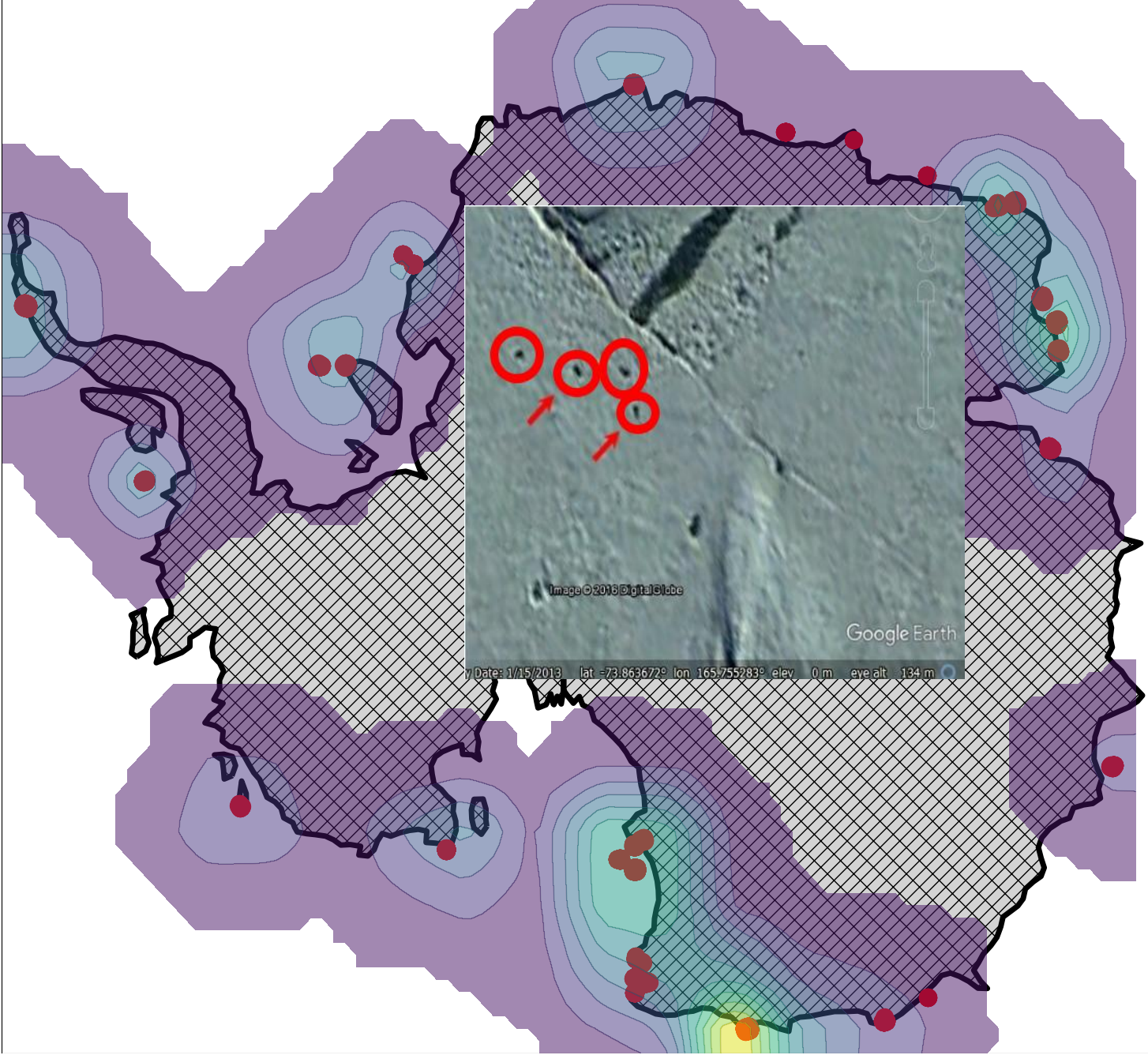}
		\caption{}
		\label{fig:seals}
	\end{subfigure}
	\begin{subfigure}[t]{0.30\textwidth}
		\centering
		\includegraphics[width = \textwidth]{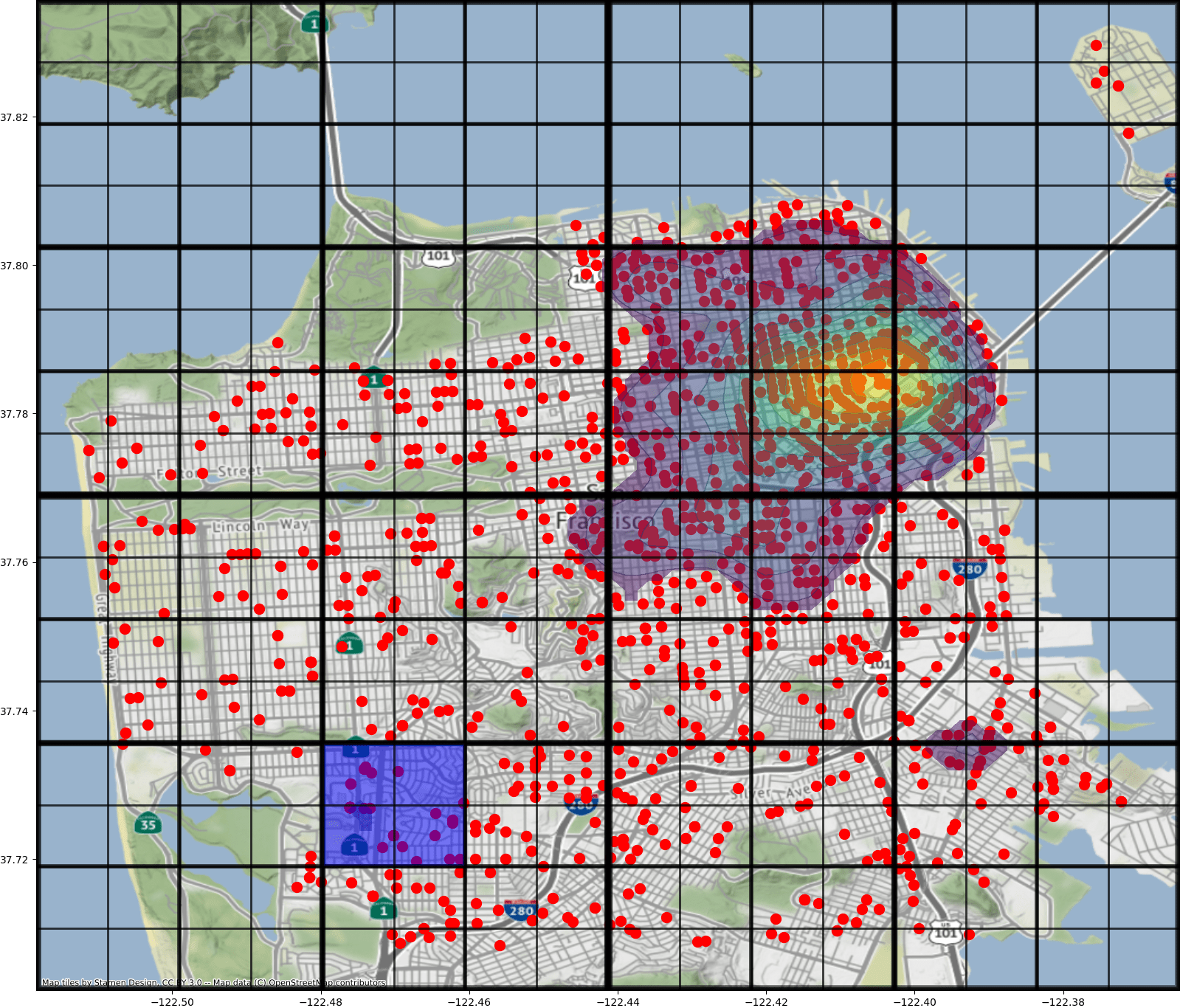}
		\caption{}
		\label{fig:sfrate}
	\end{subfigure}
	\begin{subfigure}[t]{0.30\textwidth}
	\centering
	\includegraphics[width=\textwidth]{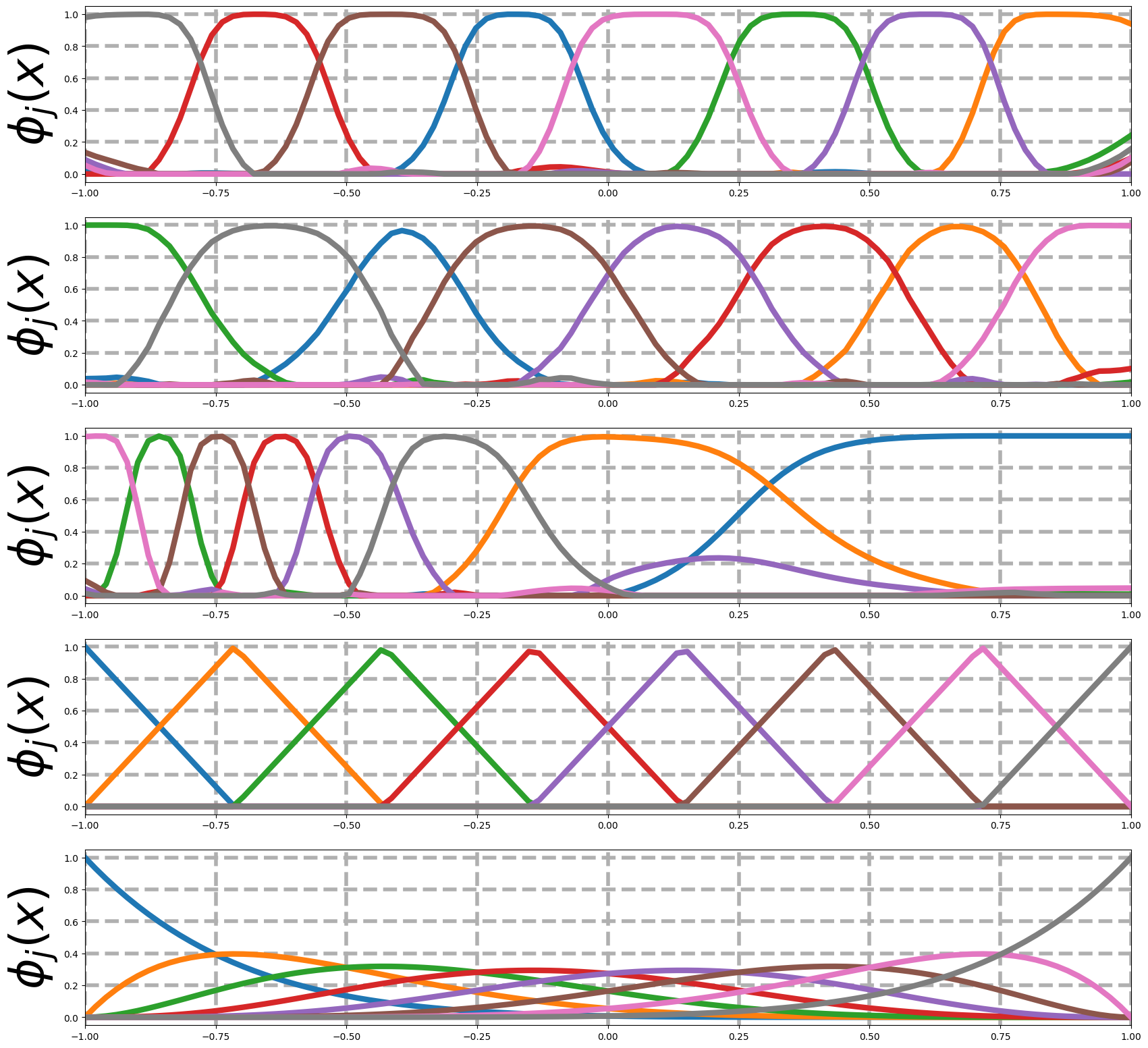}
			\caption{}
	\label{fig:triangle-basis}
\end{subfigure}

	\caption{a) The rate of occurrence of seals overlaying the map of the Antarctic. In the middle of the map we see an example of a satellite image that is used to identify Antarctic seals automatically with recognition technology [source: Google Maps]. 	
b) The estimated rate of San Francisco burglary occurrence  modeled by a Cox point process. Sensing regions are formed by hierarchical splitting (quadtree). The blue region is an example of a sensing region -- only events inside it are observed. 
	c) Positive bases (refers to Section \ref{sec:positive}), in row order:  \emph{minimal description (optimal) positive basis} for the squared exponential kernel, the Laplace kernel, the non-stationary Gibbs kernel with decreasing variation from left to right (note on the right side fewer basis functions are required); the \emph{triangle basis}; \emph{Bernstein polynomials}; all with $m = 8$.}
	\label{fig:banner}
\end{figure*}

To give a concrete example, consider the task of estimating the size of an animal population. This is a recurring task faced by ecologists, especially in the context of understanding the impact of climate change on the habitat of species. Classical approaches to this problem rely on the capture and release principle, where animals are captured, tagged, released and recaptured or tracked. These approaches are costly and time consuming, hence remote sensing from satellites or aircraft combined with recognition software are currently being investigated as promising candidates to replace some classical approaches. \citet{Guirado2019} investigate whale identification from satellite data, and \citet{Goncalves2020} count Antarctic seal population from satellite data. Modern high resolution satellite imagery enables recognition of individual Antarctic seals, as illustrated in Fig.~\ref{fig:seals}. In the same figure, one can see that the occurrence of seals can be fitted well by a smoothly varying Cox process. Suppose we seek to identify the level sets of the intensity function, i.e., locations where the intensity is above a certain threshold -- defined to be the habitat. In order to use monitoring resources or annotators' time on the images efficiently, we want to minimize the number of sensing sessions needed to estimate the quantity of interest. Satellites or aircraft can cover only a limited region at once, and need to zoom in on a specific region to facilitate identification, hence finding an efficient sequence of sensing regions is a practical problem that can be addressed by methods developed in this work. Sensing regions can be chosen, for example, as rectangular blocks that stem from the hierarchical splitting of the domain as in Figure \ref{fig:sfrate}. 


\paragraph{Contributions} \looseness -1
\textbf{1)} We formulate three adaptive sensing problems for Cox processes: maximum identification, level set estimation and maximum event capture. We propose simple and efficient algorithms for each of them based on posterior sampling and top-two posterior sampling. \textbf{2)} We address the inherent positivity constraint of the intensity function by employing representations that naturally allow to enforce the positivity constraint. To this end, we introduce a novel \emph{minimal description positive basis} problem that, suitably relaxed, reduces to non-negative matrix factorization. We further relate the minimal description positive basis to other positive bases from prior work, and demonstrate their application to non-stationary kernels. \textbf{3)} Our algorithms are easy to implement and rely solely on convex optimization methods. We show that they outperform algorithms based on optimism, Laplace approximation and Bayesian V-optimal experiment design, and demonstrate their applicability to real-life problems. 

\section{PROBLEM STATEMENT} \looseness -1
Let $\mD \subset \mR^d$ be a compact domain. A \emph{Poisson point process} $\mathfrak{P}$ is a random process such that for any subset $A \subseteq \mD$, $N(A)$ -- the random variable denoting the number of events in $A$ -- is distributed as $N(A) \sim \text{Poisson}\left(\int_{A} \lambda(x) dx \right)$, where $\lambda(x)$ is a positive \emph{intensity function}. The lowercase $n(A)$ denotes the realizations of random variables $N(A)$. If $A,B\subset \mD$ and $A \cap B = \emptyset$ then $N(A)$ and $N(B)$ are independent given $\lambda$. If, in addition $\lambda(x)$ is modeled as a random function such as a truncated Gaussian process, the combined structure is referred to as \emph{doubly stochastic} or \emph{Cox process} \citep{Kingmann2002, Snyder2012}.

\looseness -1 When we observe events $\{x_j\}_{j=1}^{n(A)}$ in a sensing region $A\subset \mD$ for duration $\Delta$, their likelihood given the intensity function $\lambda$ can be calculated as  $\pP((A,\{x_{j}\}_{j=1}^{n(A)}) \propto$
\[ \exp \left( - \Delta \int_{A}\lambda(x)dx  \right)\prod_{j=1}^{n(A)} \frac{\lambda(x_{j})^{n_{x_j}}}{n_{x_j}!}.\] To simplify the exposition, we assume without loss of generality that all events have distinct locations (no repetitions), hence $n_{x_j} = 1$ for all $x_j$. This likelihood, together with a truncated Gaussian process prior allows, via Bayes' theorem, to obtain a posterior distribution of $\lambda$. Compressing the notation of acquired data to $D =(A,\{x_{j}\}_{j=1}^{n(A)})$, $P(\lambda | D) = \frac{p(D|\lambda )p(\lambda)}{\int p(D|\lambda')p(\lambda')d\lambda'}$, where the denominator contains a double integral -- sometimes referred to as \emph{double-intractability} \citep{John2018}. Two core issues with this posterior inference are that the prior $p(\lambda)$ needs to be restricted to positive Gaussian processes - itself a non-trivial task, and the double intractability needs to be efficiently handled. In Section
~\ref{sec:positive}, we show how representation in a specific basis solves these problems, but first we state our sensing problems. 

\subsection{Sensing Problems and Protocol}\looseness=-1
We consider the problem of adaptive sensing of Poisson processes. We sequentially pick {\em sensing regions}, observe events inside them at some {\em cost}, with the goal to achieve one of the following three {\em objectives}:
\begin{itemize}\looseness=-1
	\item\looseness=-1 \emph{Event capture maximization}: Capture as many events of $\mathfrak{P}$ as possible;
	\item\looseness=-1 \emph{Estimation of level sets:} Given a threshold $\tau$,  find the largest $S \subseteq \mD$ s.t. $\lambda(x) \geq \tau$ $\forall x\in S$;
	\item\looseness=-1 \emph{Maximum identification:} Identify (approximately) $x \in \arg\max_{x\in \mD}\lambda(x)$.
\end{itemize}

In all these tasks, we aim to minimize the cost of sensing. While sensing the whole domain might be optimal in terms of information provided, it is often practically infeasible due to high cost. Hence either restricting the allowed sensing regions to small regions of the domain or introducing the notion of costs is necessary. 
\paragraph{Sensing Actions}\looseness -1
We acquire information about a subset of the domain $\mD$ for a certain duration $\Delta$, for example by zooming in with a satellite. We refer to such a sensing event as an \emph{action}. The set of actions is parametrized by a collection of non-empty 	sets $\mA$, where $|\mA| < \infty$. In each iteration, we pick a sensing action $A \in \mA$ that we sense for a duration $\Delta$, and then receive locations of events $\{x_i\}_{i=1}^{n(A)}$ in $A$ only.

Each action $A$ has an associated {\em cost} calculated as $w_t(A, \Delta)$, where the cost function $w_t: \mA \rightarrow \mR$ is known. 
In this work, we assume the cost is time invariant and proportional to the volume (area) of the sensed set $w(A,\Delta) \propto |A|\Delta$, where $\Delta$ is a fixed minimum sensing time duration. Longer sensing is possible by repeating this action sufficiently often. The formalism we develop, however, is more general and can be applied with other cost models. In particular, we focus on two cost functions: \emph{uniform costs} $w(A)=C_1|A|$ and \emph{fixed costs}, where a $w(A) = C_1|A|+C_2$ for $C_1, C_2\geq0$ which introduces the preference to sense larger sets, since with uniform costs, sensing smaller parts of the set is always preferred over sensing a larger set containing them, which might be sometimes undesired.
\paragraph{Protocol}  \looseness -1 The sensing algorithms that we develop follow a simple meta-protocol consisting of three steps. In the first, we collect all information gathered so far, then pick a new sensing region as a solution to $A_t = \arg\max_{A\in \mA} Q_t(A)$, where we refer to $Q_t$ as the \emph{acquisition function}, which represents the utility of the next action and is objective dependent. In the last step, we sense $A_t$ to observe events occurring inside it. We repeat the procedure until our budget is exhausted or we are sufficiently satisfied in the progress. We can see this protocol in Algorithm \ref{alg:capture-ucb} in steps 2, 3 and 4. 
\subsection{Modeling Assumptions} \looseness -1
\paragraph{Truncated Gaussian Process}\looseness -1 We put a prior on the intensity function $\lambda$ and assume that it can be modeled as a sample from a truncated Gaussian process. By truncated Gaussian process, we understand a zero-mean Gaussian process with kernel $k$ restricted such that its samples are functions bounded away from zero. Namely, we assume that $ 0 < l \leq \lambda(x)$ for all $x \in \mD$ and $l$ known. Generating samples can be implemented using rejection sampling. The prior is proportional to a GP prior with the truncation entering via multiplication by a functional indicating whether $\lambda \geq 0$ everywhere in the domain (see \citet{Rasmussen2006}[sec. 6.2.3] for specifying GP priors with functionals).

\paragraph{Poisson Process Likelihood}  \looseness -1
Suppose that there are $t$ sensing sessions during which we sensed regions $A_j$ for $j \in [t]$, and we observed locations of the events $\{x_{j,i}\}_{i=1}^{n(A_j)}$. We use the Poisson point process likelihood, Bayes' rule and take the logarithm at the end to arrive at the following maximum a posteriori (MAP) estimation $\hat{\lambda}= \argmax_{\lambda \in \mH_k, \lambda \geq 0} U(\lambda)$, where
\begin{equation}\label{eq:potential} 
	U(\lambda) = \sum_{j=1}^{t} \sum_{i=1}^{n(A_j)} \log(\lambda(x_{j,i}))\\ -\int_{A_j} \lambda(x)dx - \frac{\norm{\lambda}_{k}^2}{2}.
\end{equation}
The inequality $\lambda \geq 0$ denotes the positivity constraint, $\mH_k$ is the {\em reproducing kernel Hilbert space (RKHS)} associated with kernel $k$, and $\norm{\lambda}_k$ is the associated norm. This expression is possible due the well-known connection between RKHS functions and MAP of Gaussian Processes (for details see \citet{Kanagawa2018}). The functional $U(\lambda)$ in the minimization problem is commonly referred to as energy, and the posterior distribution is proportional to $\exp(-U(\lambda))$. 
\paragraph{Outline} \looseness -1
There are two main challenges that we need to address to formulate sensing algorithms. First we need to incorporate the positivity constraint in the inference procedures as well as address the \emph{double intractability}. Secondly, since our algorithms rely on posterior sampling, we need to efficiently sample from a seemingly intractable posterior. We address these two issues in Section \ref{sec:positive} and \ref{sec:sampling}, respectively. Having this, we can design acquisition functions in our sensing protocol tailored to each objective separately. 
\footnote{The code can be found at: \url{https://github.com/Mojusko/sensepy}}

\begin{algorithm}
	\caption{\textsc{Cox-Thompson} and \textsc{Top2} Algorithms}
	\label{alg:capture-ucb}
	\begin{algorithmic}[1]
		\Require Kernel $k$, total cost $C$, cost function $w$, $\varphi_A = \int_A\phi(x)dx$,  $ t = 1$.
		
		\While {$\sum_{j=1}^t w(A_t) \leq C$} 
		\State Update $U(\theta)$ for sampling $\tilde{\theta} \sim \exp(-U(\theta))$
		\State 	$A_t = \begin{cases} \arg\max_{A \in \mA} \varphi_A^\top \tilde{\theta}
		  & \text{  \textsc{Cox-Thompson}}	\\
			A_t \text{ as in } \eqref{eq:top2-acq}  & \text{  \textsc{Top2}} 
		\end{cases}$
		
		\State 	Sense $A_t$, and receive
		$(n(A_t),\{x_i\}_{i=1}^{n(A_t)})$ 		
		\State $t = t +1 $
		\EndWhile
		
	\end{algorithmic}

\end{algorithm}

\section{POSITIVE BASES}\label{sec:positive} \looseness -1
One of the core challenges for inference in our model compared to classical Gaussian process (GP) regression is the positivity of the intensity function ($\lambda \geq 0$). The global nature of the constraint hinders the application of the representer theorem to make optimization tractable. Practical approaches addressing this positivity constraint are threefold: one could either employ a link function that transforms the output space to positive values; discretize the space; or choose a so called \emph{positive basis}. All approaches have their pros and cons that we discuss at the end of this section and provide references. In this work, we use a technique that falls into the positive basis approach.

By a positive basis we mean that the intensity can be approximately represented as $\lambda(x) \approx \sum_{i=1}^{m}\phi_i(x)\theta_i$ via nonnegative basis functions $\phi_i(x)\geq 0 ~ \forall x \in \mD$. This way, the positivity constraint can be implemented simply by enforcing that the inferred $\theta_i \geq 0 ~ \forall i \in [m]$. Additionally, the integral in the likelihood can be succinctly represented as $\varphi_A := \int_{A}\phi(x)dx$ and pre-computed, eliminating the double intractability.

We use this representation for the \emph{prior}, and specify the necessary distribution over $\theta$ such that it approximates the truncated GP prior. As a GP is fully specified by its mean and covariance, we need to make sure the approximated GP matches them. The core challenge is to match the covariance of the approximating GP, where $\hat{f}$ are the samples, i.e., ensure ($\stackrel{!}=$)
\[ \mE[\hat{f}(x)\hat{f}(y)] = \phi(x)^\top \mE[\theta \theta^\top ] \phi(y) \stackrel{!}=  k(x,y) ~ \forall x,y \in \mD. \]
\looseness -1 Unfortunately, this constraint cannot be satisfied globally due to the finite number of basis functions, which have limited representational power. However, it can enforced on a set of representative nodes $t_i \in \mD, i \in [m]$, which ensure it is approximately satisfied in the rest of $\mD$. Evaluating at nodes, the covariance of the approximate Gaussian prior needs to be equal to $\bGamma^2 = \bV^{-1}\bK\bV^{-1}$, where $\bV_{ij} = \phi_i(t_j)$ is the change of basis matrix, and $\bK_{ij} = k(t_i,t_j)$ the kernel matrix evaluated at the nodes. The approximation at the nodes $t_i$ is exact, while in between the nodes $t_i$, the value of the approximate covariance is interpolated. To simplify notation we use the basis $\Phi(x) = \bGamma \phi(x)$, where the positivity constraint becomes $\bGamma\theta \geq 0$ and the Gaussian distribution in the prior is  $\theta \sim \mN(0,\bI_m)$. The nodes $t_i$ are selected such that the matrix $\bV$ is invertible, and the approximation error is low. In the next section we provide  a concrete example of such a basis. However, generally, nodes are selected as equally spaced grids \citep{Papp2014} (generalized to arbitrary dimensions via Cartesian products). Note that the approximation error decreases with the size of the basis $m$.

Using the above representation, the energy in \eqref{eq:potential} and the MAP estimate becomes a convex function and a simple optimization problem, respectively:
\begin{equation*}\label{eq:penalized-likelihood-triangles}
	\hat{\theta} = \argmax_{\theta \in \mR^m, \bGamma \theta \geq 0 } \sum_{j=1}^{t} \sum_{i=1}^{n(A_j)}\log(\theta^\top \
	\Phi(x_i)) - \theta^\top \bGamma \varphi_A - \frac{1}{2} \norm{\theta}_2^2,
\end{equation*}
where $\varphi_A = \int_{A}\phi(x) dx$. The optimization problem can be efficiently solved to near-optimality using interior point methods \citep{Nemirovski2008}. Now we study how to choose the basis $\phi(\cdot)$ such that the above construction has low approximation error.
\subsection{Minimal Description Positive Basis}\looseness -1
We want to select $\phi(\cdot)$ as the {\em minimal description positive basis}. By that we mean a basis, which has the least average squared error for a fixed size $m$. This is desirable, since the computational burden of approximate inference is proportional to complexity of matrix-vector multiplication of sizes $m$, i.e., $\mO(m^2)$ basic linear algebra operations.

\looseness -1 Let us represent the evaluation map as an operator $L = (\phi_1(\cdot) \dots \phi_m(\cdot))$ from $\mR^m_+ \rightarrow  L_2(\mD)$. The minimal description positive basis, also called \emph{optimal basis} with size $m$, can is defined as the solution to the following optimization:
\[ L^* = \argmin_{\substack{L:\mR^m_+ \rightarrow L_2(\mD),~ \norm{Le_i}_2 = 1 \\ \braket{\delta(x),Le_i}_2  \geq 0  ~\forall x \in \mD}} \int_{\substack{f \geq 0}}  \min_{y \in \mR^k_+}\norm{f - Ly}_2^2 dF(f).\]
\looseness -1 Let us unpack the meaning of this optimization problem. Moving from inner to outer problems, we first look for a positive combination of basis functions, which has lowest $L_2(\mD)$ under the expectation over the prior of $f$. Then in the outer minimization, we search for a basis representation that minimizes the expected error, and is positive (we add normalization to make the solution well-defined). This reconstruction error minimization has the same goal as the Karhunen-Lo\'eve decomposition (stochastic version of Mercer decomposition) where the basis functions are orthogonal. It is known that the Karhunen-Lo\'eve decomposition has the lowest average $L_2(\mD)$ error among all orthonormal bases \citep{Adler2009}. Hence, our construction can be considered a positive counterpart to the Karhunen-Lo\'eve basis.

This functional optimization problem is intractable due to, among other things, the difficulty of integration. Nevertheless, we can perform two approximations: assume a finite domain $|\mD|=n$, and approximate the integral with samples to get
\begin{align*}
\argmin_{\substack{\bL \in \mR^n \times \mR^m, ~ \bL \geq 0, ~ \norm{\bL e_i}_2 = 1  }} \min_{\substack{\bY \in \mR^m\times \mR^s, ~ \bY \geq 0}} \norm{\bF - \bL \bY}_F^2. 
\end{align*}
After these relaxations, the problem remarkably becomes equivalent to non-negative matrix factorization \citep{Gillis2020} -- a challenging but heuristically solvable problem for which efficient solvers are readily available. The rows of $\bF\in \mR^{n \times s}$ are the sample paths sampled from the truncated GP, and $\bY\in \mR^{m \times s}$ incorporates the weights $y$. The evaluation operator $L$ becomes a matrix on a finite domain. 

The resulting basis can be seen in Fig. \ref{fig:triangle-basis} (1st row) for the squared exponential kernel. The effect of using a kernel that generates rougher sample paths, such as the Mat\'ern or Laplace kernel, causes the basis to be \emph{more peaked} as in Fig.~\ref{fig:triangle-basis} (2nd row). To choose the nodes $t_i$ for the optimal basis, we pick the peaks of the bumps in Fig.~\ref{fig:triangle-basis}. With this choice, we see that $\bV_{ij} = \phi_i(t_j)$ will become close to the identity matrix, as each separate basis function (single color) is zero at the peaks of other basis functions. 
\paragraph{Non-stationary kernels} \looseness -1 Of special interest are optimal positive bases for non-stationary kernels, e.g., the Gibbs kernel $k(x,y) = \exp(-(x-y)^2/(\gamma(x)^2+\gamma(y)^2))$ with varying lengthscale $\gamma$ as in Fig.~\ref{fig:triangle-basis} (3rd row). In this example, the regions of the domain with low spatial variations do not need to be approximated by large numbers of basis functions and the optimal basis adapts accordingly. This is of special importance for us, since in spatial modeling, the ultimate goal of this work, the problems often exhibit non-stationarity due to varying geographical features such as effects of water, mountains, etc. This can be modelled by what we call \emph{indicator modification}, e.g., $k(x,y) = w(x)^\top w(y)\exp(-(x-y)^2/\gamma^2)$, where $w(x)$ summarizes geographical features, and $k$ is a non-stationary kernel. We will see concrete examples in the experiments.
\subsection{Other positive bases} \looseness -1 Other positive bases used in prior works are the famed Bernstein polynomials (Fig. \ref{fig:triangle-basis}, 5th row) \citep{Papp2014}, \emph{triangle basis} (Fig. \ref{fig:triangle-basis}, 4th row)  or positive Hermite splines \citep{Alizadeh2008}. The \emph{triangle basis} of \citet{Maatouk2016} and \citet{Cressie2008} stands out. Qualitatively, it is very similar to the optimal basis (in case of stationary kernels), is well-conditioned, has good approximation properties and the integrals $\varphi_A$ can be calculated in closed form. It has also been used in prior works for positive constraints \citep{Lopez-Lopera2019}. Its choice was motivated by convenience, but we believe this work justifies its good empirical performance by its closeness to the optimal basis representation for stationary kernels. We introduce it in more detail in the next paragraph. 

\paragraph{Triangle basis} \looseness -1
Given a domain $\mD = [-1,1]$, let us define the individual triangle basis functions $\phi_j$ for $j \in [m]$:
	$\phi_j(x) = \begin{cases}
		1 - |\frac{(x-t_j)m}{2}| &   \text{if } |\frac{(x-t_j)m}{2}| \leq 1 \\
		0 &  \text{ otherwise}
	\end{cases}$.
As the basis functions are bounded $0 \leq \phi_i(x)\leq 1$ and do not overlap at node points $t_i$, the constraints $l \leq \lambda(x)$ can be represented as linear constraints on $\theta$: $l \leq \theta$  \citep[for details refer to][]{Lopez-Lopera2018}. Now, in order to use this basis with kernel $k$, we transform it by $\Phi(x) = \bGamma \phi(x) = \bV^{-1}\bK^{1/2}\phi(x)$ as above, where we note that $\bV$ is the identity when we use the same nodes $t_i$ as in the definition.

\subsection{Why not link functions?}
\looseness -1 Many prior works model the intensity function $\lambda= \mu(f)$ with a {\em link function} $\mu$ to ensure positivity. These include: sigmoid \citep{Adams2009}, exponential \citep{Moller1998} and square \citep{Flaxman2017}. The problems with link functions are twofold: the energy $U$ is either non-convex or the integral $\int \lambda(x)dx$ cannot be evaluated in closed form nor precomputed -- the so called \emph{double-intractability} -- both of which are avoided by positive bases. The square link function seems to be the most promising, as the double intractability can be alleviated with a fixed basis, but it results in non-convexity of the energy due to the equivalence of $-f$ and $f$ as solutions. It also leads to artefacts called \emph{nodal lines} \citep{John2018}, where, as the range of $f$ changes sign,  the approximation of $\lambda$ is poor (see Appendix \ref{app:nodal-lines}). The sigmoid link function leads to a convex energy, but due to the \emph{double-intractability} it requires numerical integration, which makes the computation costly as we will show.\footnote{The sensing algorithms we develop can be used with sigmoid as well; they are just not as computationally efficient.}
\section{POSTERIOR SAMPLING AND ACQUISITION FUNCTIONS}\label{sec:sampling}
\looseness -1 Sampling from the posterior over the intensity function, as introduced below, is a cornerstone for our algorithms. To sample from the posterior $P(\theta|\{A_i\}_{i=1}^t)$ using a positive basis, we need to sample from $P(\theta|D) \propto \exp(-U(\theta))$, where $U(\theta) =  \sum_{j=1}^{t} \sum_{i=1}^{n(A_j)}-\log(\theta^\top \Phi(x_{ij})) + \theta^\top \varphi_{A_j} + \frac{1}{2} \norm{\theta}_2^2$ and $\theta$ corresponds to the weights in the positive basis. Fortunately, in this representation, $U(\theta)$ is convex on $\bGamma\theta \geq 0$ or $\bGamma\theta \geq l$, when applying the lower bound, and hence the posterior distribution is {\em log-concave}. Log-concavity of a distribution is a desirable property that leads to efficient and even provably consistent approximate inference \citep{Dwivedi2018}, and can be solved, e.g., via Langevin dynamics. Langevin dynamics is a stochastic process that follows a discretized stochastic differential equation defined via a gradient flow on the energy $U$ (see \eqref{eq:langevin} below for an example). In our case, the only difficulty arises from the fact that $U$ is not smooth, as outside of $\bGamma\theta \geq l$ it is zero. Recent advances extend Langevin dynamics to \emph{non-smooth} energies as we show below.
\paragraph{Proximal Langevin Dynamics} Let us define an indicator function $g(\theta) = \mathbf{i}_{l \leq \bGamma\theta}$, taking values  $\{0,\infty\}$, and let our modified energy defined on $\mR^m$ be $\tilde{U}(\theta) = U(\theta)  + g(\theta)$. \citet{Durmus2016} and \citet{Brosse2017} propose a proximal version of Langevin dynamics utilizing the Moreau-Yosida envelope of a function $g$ \citep{Borwein2005}. While, the details of the derivation of this algorithm are out of scope of this paper, the final algorithm is remarkably simple and relies on the proximal operator $\operatorname{pr}_g(\theta) = \arg\min_{x}\left(  g(x) + \eta \norm{x - \theta}_2^2  \right)$, which for our $g$ simplifies to projection on a polytope $ \operatorname{pr}(\theta) = \min_{\bGamma y \geq  l} \norm{\theta-y}^2_2$, which can be very efficiently solved using a quadratic programming algorithm of \citet{Goldfarb1983}.

The Moreau-Yosida Unadjusted Langevin Algorithm (MYULA) of \citet{Brosse2017} is defined via the stochastic update rule,
\begin{eqnarray} \label{eq:langevin}
	\tilde{\theta}_{k+1} \gets (1-\eta)\tilde{\theta}_k- \eta \nabla U(\tilde{\theta}_{k}) + \eta \operatorname{pr}(\tilde{\theta}_k) + \sqrt{2\eta}w_{k}
\end{eqnarray}
\looseness -1 which requires step size $\eta = 1/(L+1)$, where $L$ is the Lipschitz constant of $U(\theta)$ and $w_{k} \sim \mN(0,\bI_m)$. For any $l>0$, the Lipschitz constant is bounded and can be found by approximate power method applied on the Hessian of $U$. The initial point is set to coincide with the MAP. As $t \rightarrow \infty$, the continuous limit of the above equation leads to exact posterior sampling, and hence the distribution of $\tilde{\theta}_t$ closely approximates the sample from $P(\theta)\propto \exp(-\tilde{U}(\theta))$ in TV distance \citep{Brosse2017}. 

\looseness -1 Other Langevin dynamics algorithms \citep{Hsieh2018,Zhang2020a} and other specialized approximate inference schemes \citep{Donner2018} are applicable in this context as well as we show in Appendix \ref{app:langevin}. However, MYULA seems to be suited for this problem as it can be easily rerun when new data arrives, and it can deal with ill-conditioning of $\bGamma$, which is often the case for smooth kernels. With proximal sampling, the ill-conditioning appears only within a quadratic program, which can be easily addressed via second-order methods. Other methods rely on first-order approaches, thus ill-conditioning forces step sizes to be small, leading to longer compute times.

\subsection{Maximizing Event Capture} \label{sec:ucb} \looseness -1
Using the above, we can formulate an algorithm for the first objective: \emph{maximizing event capture}. The goal here is to capture as many events subject to the cost of sensing. Each $A \in \mA$ has an associated cost, $w(A)$. We want to spend the budget in the most effective way by minimizing the following regret,
\begin{equation}\label{eq:regret} \looseness -1
	R(\{A_t\}_{t=1}^T) = \sum_{t=1}^T w(A_t)\frac{\mE[N(A^*)]}{w(A^*)} - \mE[N(A_t)],
\end{equation}
where $A^*$ maximizes $\mE[N(A)]/w(A)$ over $A \in \mA$ and expectation is over the sampling from the Cox process. The regret represents the average missed events incurred by investing the budget $\{w(A_t)\}_{t=1}^T$ to potentially suboptimal actions in terms of expected number of observed events to cost ratio. We propose to use posterior sampling to solve this task \citep{Russo2017a}. In connection to Poisson process sensing, we refer to this algorithm as \textsc{Cox-Thompson}, where we use \eqref{eq:langevin} with positive bases to sample $\tilde{\theta}$ from the posterior distribution. Afterwards, we sense $A \in \mA$ which leads to the best count to cost ratio under the sampled parameter $\tilde{\theta}$ as in Alg.~\ref{alg:capture-ucb}. This algorithm, is essentially a greedy algorithm that minimizes the regret of \eqref{eq:regret} under the current posterior distribution and is known to converge under mild conditions \citep{Russo2017a}.

\subsection{Learning Level Sets and Maxima} \label{sec:level-sets} \looseness -1 To identify the maximum of $x^* = \argmax_{x}\lambda(x)$ and level sets, we rely on the seminal re-sampling heuristic of \citet{Chernoff1959}, which was recently analyzed, and popularized by \citet{Russo2016a}. Namely, the heuristic relies on resampling from the posterior until another sample leads to a different recommendation of what is the maximizer or what are the level sets of $\lambda$. Having these two recommendations, we sense a region that leads to the decrease in uncertainty of this recommendation. In particular, the recommendation rule for maximum $x^*$ is $\hat{x}_1 = \arg\max_{x\in \mD} \tilde{\lambda}(x)$, where $\tilde{\lambda}(x)$ is the sample from the posterior. Then, we re-sample from posterior until we find a different recommendation $\hat{x}_1\neq \hat{x}_2$. After that, we randomly choose to sense \emph{the lowest cost sets $A_1$ and $A_2$ containing $\hat{x}_1$ and $\hat{x}_2$, respectively} with equal probability. We refer to this algorithm as \textsc{Top2}.  This algorithm is known to be consistent for any uncorrelated exponential family random responses, which includes Poisson responses, as shown by \citet{Russo2016a}. In our case, we exploit correlation structure of the posterior, which we believe can  only improve the efficiency of this scheme. 

In order to implement an algorithm that identifies level sets of $\lambda(x)$, in other words, the largest $S\subset \mD$ s.t. $\lambda(x) \geq \tau$ for all $x \in S$, we resample until the prediction of level sets $\tilde{S}_1 \neq \tilde{S}_2$ are different. The difference in recommendation is perhaps best summarized by an exclusive OR (XOR) operation on the two sets $Z = \tilde{S}_1 \oplus \tilde{S}_2$, and then we choose an action $A$ that shares the most overlap with the resultant set $Z$. We weight the overlap by the respective difference in magnitude of the two random samples over $Z$, namely, using the positive bases where two samples are $\tilde{\theta}_1$ and $\tilde{\theta}_2$,
\begin{equation}\label{eq:top2-acq}
	A_t = \argmax_{A \in \mA} \int_{A} \frac{|\Phi(x)^\top(\tilde{\theta}_1 - \tilde{\theta}_2)|}{w(A)} (\tilde{S}_1  \oplus \tilde{S}_2)(x)dx.
\end{equation}

\begin{figure*}
	\includegraphics[width=\textwidth]{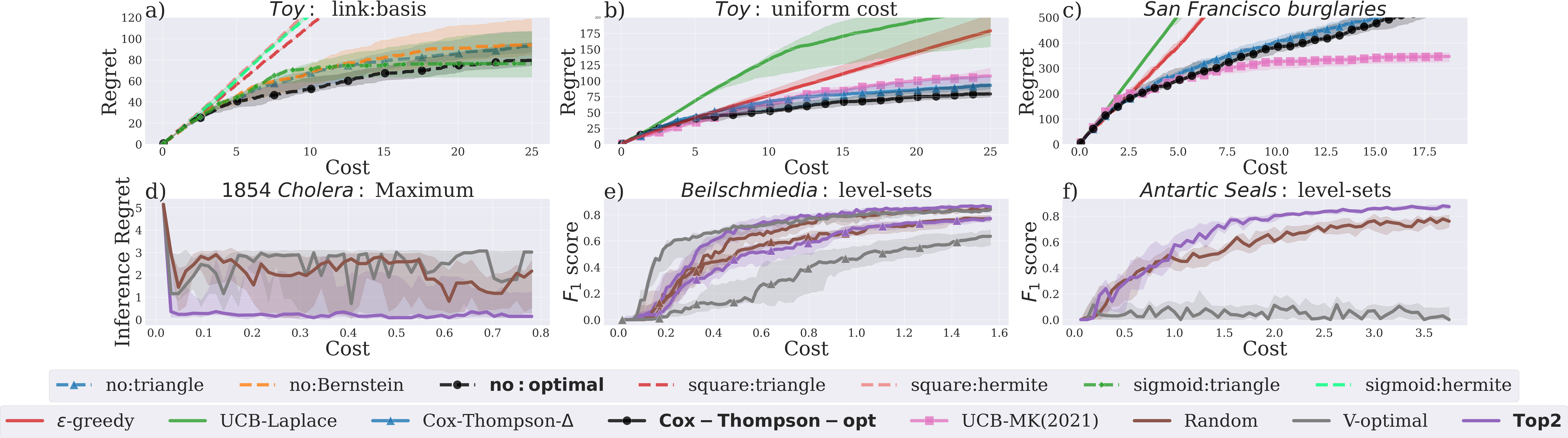}
	\caption{Experimental results: 10 repetitions with 25\%, 50\% and 75\% quantiles in shaded regions. Our main algorithms are in bold font. In a), we analyze performance due to different positive basis and link functions. Our legend uses the format `link:basis', where 'no' corresponds to no link function. In b) and c), we study maximum capture of events; in c), specifically,   burglaries fitted on the \emph{San Francisco} dataset. In d), we analyze maximum identification on the \emph{cholera} dataset. Inference regret (defined in the text) measures suboptimality in this task. In e) and f), we study habitat inference, i.e., classification of level sets as below a threshold or above, evaluated using $F_1$ score (prediction with MAP estimate always). The triangle markers refer to using the triangle basis instead of the optimal basis (dots) with the same number of basis functions, which in b), c) and e) lead to worse results. Overall, our \textsc{Top-2} algorithm outperforms other approaches, and our \textsc{Cox-Thompson} is top performing along with \textsc{UCB-MK}.}
	\label{fig:experiments}
\end{figure*}

\section{DISCUSSION \& EXPERIMENTS}\looseness=-1
Before we proceed to compare the sequential algorithms, let us first investigate how positive bases compare with link functions in two metrics: validity to solve the task and execution time. We study event capture maximization on a \emph{toy example} function $\lambda(x) = 4\exp(-(x+1))\sin(2x\pi)^2$ on $[-1,1]$ in Fig.~\hyperref[fig:experiments]{3a}, which is well modeled by the squared exponential kernel. Our experiments with link function are done with a finite basis as well, because performing inference for each $\lambda(x)$ for each $x \in \mD$ is computationally infeasible. 

We see that the square link function fails to solve the task due to nodal lines (Appendix \ref{app:nodal-lines}). Sigmoid link functions with Quadrature Fourier Features (QFF) \citep{Mutny2018b}, a minimal description basis (non-positive), fails because the features $\Phi(x)$ are difficult to integrate. On the other hand, using the sigmoid link with the triangle basis performs just as good as the constrained version -- however it is orders of magnitude slower (cf., Fig.~\ref{fig:extra} in Appendix \ref{app:experiments}). Lastly, using the optimal basis performs essentially the same as the triangle basis, because they are indeed very similar on this stationary problem. However, the difference appears when using non-stationary kernels as we will see in other applications.
\paragraph{Baseline Sensing Algorithms}\looseness -1
For the count-regret minimization task, we compare our approach with the UCB algorithm \citep{Srinivas2009} based on credible sets from the Laplace approximation (\textsc{UCB-Laplace}), the UCB algorithm of \citet{Mutny2021a} (\textsc{UCB-MK}), which uses a regression estimate instead of MAP with stylized confidence parameter $\beta=3$, and \emph{$\epsilon$-greedy} with decreasing exploration probability. 
For the level-set and maxima identification, we compare against random search and classical approximate sequential Bayesian V-optimal design \citep{Chaloner1995}.

\paragraph{V-optimal experimental design}\looseness -1
The literature on Bayesian experimental design \citep{Chaloner1995,Gotovos2013} proposes a general recipe to solve the level set estimation problem. Namely, it suggests to sense regions that reduce the uncertainty the most under the posterior expectation in the \emph{region of interest} subject to the sensing cost. The \emph{region of interest} $R_t$ depends on our objective and is time varying. If learning levels sets above a threshold $\tau$, then $R_t = \{ x \in \mD| \operatorname{lcb}_t(x)\geq \tau  \}$, while for identifying maxima, $R_t = \{ x \in \mD | \operatorname{ucb}_t(x) \geq \max_{z\in \mD}\operatorname{lcb}_t(z)\}$. As we proceed with iterations, the region of interest shrinks to the smallest such set. Following the derivation of Bayesian experimental design, we greedily select the next sensing region $A \in \mA$ to maximally reduce the posterior squared prediction error in $R_t$  in expectation over the realization of $A$ and in expectation over the posterior of $\theta$. Such expectation is unfortunately not feasible. If we again use Laplace approximation and the principle of optimism, we can derive an acquisition function (cf., Appendix \ref{app:v-design}). We refer to this algorithm as \textsc{V-optimal} design due to its relation to Bayesian V-optimal design. This is, to our knowledge, a novel instantiation of the Bayesian experimental design framework, but it stems from well-established Bayesian experimental design principles. We explain it in more detail in Appendix \ref{app:v-design}.

\subsection{Discussion: Algorithm comparison} \looseness -1
We compare our proposed algorithms on four problems, including an additional experiment in Appendix  \ref{app:experiments} with fixed costs (concave). For the experiments presented below, we use a uniform cost function $w(A)=|A|$, hence only the smallest sets in $\mA$ (hierarchically generated) are sensed. More details and hyperparameters along with inferred intensities can be found in Appendix~\ref{app:experiments}.

The main overall message from the benchmarks is that algorithms that do not exploit correlation stemming from the kernel $k$, such as random sampling or $\epsilon$-greedy, perform worse than  algorithms that do. Secondly, algorithms based on Laplace approximation; \textsc{UCB-Laplace} and \textsc{V-optimal} fail when the rate function $\lambda(x)$ is small in some regions ($l \approx 0$), as explained in Appendix~\ref{app:ucb}. Lastly, the UCB algorithm of \citet{Mutny2021a}, which involves intricate variance rescaling, performs similarly to our very simple \textsc{Cox-Thompson} algorithm for count-regret minimization.

Our first problem in Fig.~\hyperref[fig:experiments]{2b} has the same rate function as in \hyperref[fig:experiments]{2a}, but we vary sensing algorithms instead of representations. We see that Thomspon sampling performs slightly better to the optimally tuned UCB algorithm of \citet{Mutny2021a}. \textsc{Cox-Thompson} required no tuning, and no confidence parameter. Before discussing the four real-world benchmarks individually, we provide a cautionary note regarding the modeling aspect.

\paragraph{Simplifications} \looseness -1 The experiments presented below are motivated by real problems, but significantly simplified. Each application involves a lot of additional technological, ethical and economical details that we do not consider here. Therefore, one should see the following as stylized illustrations, rather than complete solutions. In addition, some of the benchmarks can be improved by expert designed kernels $k$. For example, ecologists might have prior knowledge regarding the seals' habitat based on geographical features, and this should be incorporated into $k$ to improve efficiency. Without such expert knowledge, we rely on naive geographical features and spatial correlation. However, the strength of our formalism lies in its flexibility. 

Secondly, we want to evaluate the applicability of our approach without interference of potential model misspecification, and at the same time be able to include the randomness of the process into our algorithmic comparison. Therefore, we execute the experiments in a \emph{fit and test} approach. We first fit the model to the whole data corresponding to our application, which results in a ground truth rate function $\lambda$, and then apply our algorithms to estimated $\lambda$.

\subsection{Individual Benchmarks}
\paragraph{Burglary surveillance} \looseness=-1 In this problem, we estimate the intensity of burglary occurrences as a function over space in the city of San Francisco, as illustrated in Fig.~\ref{fig:sfrate}. The sensing duration $\Delta$ is $30$ days corresponding to, e.g., deploying mobile cameras in rectangular city blocks. We compare the regret (as in \eqref{eq:regret}) of the baselines and see that \textsc{Cox-Thompson} outperforms other algorithms in  Fig.~\hyperref[fig:experiments]{2c} except for UCB (with optimal basis) which uses a tuned confidence value $\beta$. If it was run according to the theoretical specifications, it would not be competitive. The optimal basis performs much better than the \emph{triangle} basis (blue), since the covariance of the problem is non-stationary. Namely, in parks and water areas, the rate is zero, and the optimal basis does not cover these areas with basis functions. With the same basis size, the triangle basis cannot capture the problem sufficiently accurately.
\paragraph{Cholera epidemics}\looseness=-1 In this experiment, we consider the 1854 London outbreak of cholera using John Snow's famous dataset. The goal here is to identify the pollutant's source (polluted well). We assume, as commonly done, that the rate function decays smoothly with the distance to the sources (wells). Hence, the goal is to identify the maximum of $\lambda$. To measure progress, in  Fig.~\hyperref[fig:experiments]{2c}, we report \emph{inference regret} on the $y$-axis, which measures the suboptimality of the current belief about the optimum, defined as $\max_{x\in \mD}\lambda(x) - \lambda(\hat{x}^*_t)$, where $\hat{x}^*_t$ is the current belief for the value at optimality. The simple \textsc{Top2} approach performs very well. The sensing duration $\Delta$ is set to $1/20$ of the length of the epidemic. 
\paragraph{Habitat monitoring}\looseness=-1 Suppose we want to perform surveillance of a region in order to estimate the rate of occurrence of a particular species. In order to model this application, we use occurrence data of \emph{Beilschmiedia}, a tree genus native to Africa, from \citet{Baddeley2015} and seal observations in Antarctica from \citet{Goncalves2020}, specifically a section of Antarctic territory around the Ross sea. In both cases, we model occurrence as Poisson point process and seek to determine the region where the occurrence rate is above a given threshold value $\tau$, which we take as the definition of a habitat (chosen to be half the maximum rate value). For the Antartic seals, we use an indicator kernel, which indicates whether the region is sufficiently close to the shore, and for \emph{Beilschmiedia} we use squared exponential kernel that takes the slope $s_{x,y}$ and height $h_{x,y}$ of a point $(x,y)$ as inputs, making it non-stationary in $x,y$.

\looseness -1 As the task essentially corresponds to a classification of regions according to whether the rate is above or below $\tau$, we report the $F_1$ score over the sensing rounds. The $F_1$ score is a commonly used classification metric incorporating  precision and recall jointly. Hereby, below the level is associated with label $(-)$ and above $\tau$, $(+)$, in Figs. \hyperref[fig:experiments]{2e} and \hyperref[fig:experiments]{2f}. We see that \textsc{Top2} outperforms random sampling and \textsc{V-optimal} in both instances, as it identifies the level sets more efficiently. It seems that \textsc{V-optimal} gets stuck in regions with small values $l = 0$, which are present in the \emph{seal} problem. Also, note that the triangle marked lines in Fig. \hyperref[fig:experiments]{2e} use the triangle basis approximation with the same basis size. With the same basis size $m$ as the optimal basis, it fails to approximate the function sufficiently well to solve this task.
\section{RELATED WORK}\label{sec:related} \looseness -1 

Previous works on Poisson point process intensity estimation can be roughly categorized into those based on smoothing kernels \citep{Berman1989}, and those based on positive definite kernels. Unlike the methods based on smoothing kernels, kernelized estimators need to address the positivity constraint of the intensity function. Positivity, as one of the simplest \emph{shape constraints}, can be enforced by representing our function in a specially constructed positive basis such as positive polynomials or inherently positive bases \citep{Papp2014,Maatouk2016,Lopez-Lopera2018}. \citet{Lopez-Lopera2019} and \citet{Alizadeh2008} use positive bases for inhomogenous point process inference with passively collected data, similarly as done in this work. Link functions are another way to enforce positivity, investigated by \citet{Adams2009}, \citet{Moller1998} and \citet{John2018}.

Confidence or credibility estimates are considered by \citet{Moller2003} who propose to bootstrap confidence bounds from observational data. \citet{Walder2017} use a Laplace approximation, where the intensity is modeled as square of a Gaussian process, which we exploited in our baseline algorithms. 
Adaptive sensing of Poisson point processes with sensing costs is introduced by \citet{Grant2020} and the kernelized version by \citet{Mutny2021a}. \citet{Grant2020} partition the domain and sequentially refine histogram estimators with truncated gamma priors. Their adaptive discretization is done at a specific rate in order to identify the maximum. This approach is suboptimal to approaches utilizing the correlation between bins as here. \citet{Mutny2021a} introduce adaptive sensing of Poisson point processes where the rate belongs to a reproducing kernel Hilbert space. They study the same regret as here from the frequentist perspective, however their algorithm cannot be applied to level-set nor maximum identification and their algorithm is more complicated. In broader perspective, the valuable item discovery of \citet{Vanchinathan2015} can address the capture of events with counts that have subGaussian distributions.
\section{CONCLUSION}
We studied adaptive sensing of Cox processes. We proposed three sensing objectives: learning level sets, maximum identification and event capture, and provided adaptive algorithms based on posterior sampling that optimize these objectives. In order to tackle the positivity constraint, we introduced the concept of minimal description positive basis to approximate the rate function to facilitate accurate and fast inference. We believe our algorithmic techniques can address complex real-world sensing problems and enable resource-efficient solutions. 

\newpage


\section*{Acknowledgement}
This project has received funding Swiss National Science Foundation through NFP75 and this publication was created as part of NCCR Catalysis (grant number 180544), a National Centre of Competence in Research funded by the Swiss National Science Foundation.

Mojm\'ir Mutn\'y would like to thank Pierre-Cyril Aubin-Frankowski, Huang Liaowang, Ya-Ping Hsieh and Elvis Nava for valuable discussions and suggestions.

\bibliography{master.bib}

\begin{thebibliography}{}

\bibitem[Adams et~al., 2009]{Adams2009}
Adams, R.~P., Murray, I., and MacKay, D.~J. (2009).
\newblock Tractable nonparametric bayesian inference in poisson processes with
  gaussian process intensities.
\newblock In {\em Proceedings of the 26th Annual International Conference on
  Machine Learning}, pages 9--16.

\bibitem[Adler and Taylor, 2009]{Adler2009}
Adler, R.~J. and Taylor, J.~E. (2009).
\newblock {\em Random fields and geometry}.
\newblock Springer Science \& Business Media.

\bibitem[Alizadeh et~al., 2008]{Alizadeh2008}
Alizadeh, F., Eckstein, J., Noyan, N., and Rudolf, G. (2008).
\newblock Arrival rate approximation by nonnegative cubic splines.
\newblock {\em Operations Research}, 56(1):140--156.

\bibitem[Auer, 2002]{Auer2002}
Auer, P. (2002).
\newblock Using confidence bounds for exploitation-exploration trade-offs.
\newblock {\em Journal of Machine Learning Research}, 3(Nov):397--422.

\bibitem[Baddeley et~al., 2015]{Baddeley2015}
Baddeley, A., E., R., and R., T. (2015).
\newblock {\em Spatial Point Patterns: Methodology and Applications with R}.
\newblock Chapman and Hall/CRC Press.

\bibitem[Berman and Diggle, 1989]{Berman1989}
Berman, M. and Diggle, P. (1989).
\newblock Estimating weighted integrals of the second-order intensity of a
  spatial point process.
\newblock {\em Journal of the Royal Statistical Society: Series B
  (Methodological)}, 51(1):81--92.

\bibitem[Borwein and Zhu, 2005]{Borwein2005}
Borwein, J. and Zhu, Q. (2005).
\newblock {\em Techniques of Variational Analysis}.
\newblock Springer-Verlag New York.

\bibitem[Boyd and Vandenberghe, 2004]{Boyd2004}
Boyd, S. and Vandenberghe, L. (2004).
\newblock {\em Convex optimization}.
\newblock Cambridge university press.

\bibitem[Brosse et~al., 2017]{Brosse2017}
Brosse, N., Durmus, A., Moulines, E., and Pereyra, M. (2017).
\newblock Sampling from a log-concave distribution with compact support with
  proximal langevin monte carlo.
\newblock In Kale, S. and Shamir, O., editors, {\em Proceedings of the 2017
  Conference on Learning Theory}, volume~65 of {\em Proceedings of Machine
  Learning Research}, pages 319--342. PMLR.

\bibitem[Chaloner and Verdinelli, 1995]{Chaloner1995}
Chaloner, K. and Verdinelli, I. (1995).
\newblock Bayesian experimental design: A review.
\newblock {\em Statist. Sci.}, 10(3):273--304.

\bibitem[Chernoff, 1959]{Chernoff1959}
Chernoff, H. (1959).
\newblock Sequential design of experiments.
\newblock {\em Annals of Mathematical Statistics,}, 30(30):755--770.

\bibitem[Cox, 1955]{Cox1955}
Cox, D.~R. (1955).
\newblock Some statistical methods connected with series of events.
\newblock {\em Journal of the Royal Statistical Society: Series B
  (Methodological)}, 17(2):129--157.

\bibitem[Cressie and Johannesson, 2008]{Cressie2008}
Cressie, N. and Johannesson, G. (2008).
\newblock Fixed rank kriging for very large spatial data sets.
\newblock {\em J. R. Statist. Soc. B}, 70(1):209--226.

\bibitem[Davis and Rabinowitz, 2007]{Davis2007}
Davis, P.~J. and Rabinowitz, P. (2007).
\newblock {\em Methods of numerical integration}.
\newblock Courier Corporation.

\bibitem[Diggle, 2013]{Diggle2013b}
Diggle, P. (2013).
\newblock {\em Statistical analysis of spatial point patterns}.
\newblock CRC Press (Chapman Hall Book).

\bibitem[Diggle et~al., 2013]{Diggle2013}
Diggle, P.~J., Moraga, P., Rowlingson, B., Taylor, B.~M., et~al. (2013).
\newblock Spatial and spatio-temporal log-gaussian cox processes: extending the
  geostatistical paradigm.
\newblock {\em Statistical Science}, 28(4):542--563.

\bibitem[Donner and Opper, 2018]{Donner2018}
Donner, C. and Opper, M. (2018).
\newblock Efficient bayesian inference of sigmoidal gaussian cox processes.
\newblock {\em Journal of Machine Learning Research, year 2018, volume
  19,number 67, pages 1-34}.

\bibitem[Durmus et~al., 2018]{Durmus2016}
Durmus, A., Moulines, E., and Pereyra, M. (2018).
\newblock Efficient bayesian computation by proximal markov chain monte carlo:
  when langevin meets moreau.
\newblock {\em SIAM Journal on Imaging Sciences}, 11(11):473--506.

\bibitem[Dwivedi et~al., 2018]{Dwivedi2018}
Dwivedi, R., Chen, Y., Wainwright, M.~J., and Yu, B. (2018).
\newblock Log-concave sampling: Metropolis-hastings algorithms are fast!
\newblock In {\em Conference on Learning Theory}, pages 793--797.

\bibitem[Flaxman et~al., 2017]{Flaxman2017}
Flaxman, S., Yee, W.~T., and Sejdinovic, D. (2017).
\newblock Poisson intensity estimation with reproducing kernels.
\newblock {\em AISTATS 2017}, pages 5081--5104.

\bibitem[Gillis, 2020]{Gillis2020}
Gillis, N. (2020).
\newblock {\em Nonnegative Matrix Factorization}.
\newblock Siam.

\bibitem[Goldfarb and Idnani, 1983]{Goldfarb1983}
Goldfarb, D. and Idnani, A. (1983).
\newblock A numerically stable dual method for solving strictly convex
  quadratic programs.
\newblock {\em Mathematical programming}, 27(1):1--33.

\bibitem[Gon{\c{c}}alves et~al., 2020]{Goncalves2020}
Gon{\c{c}}alves, B.~C., Spitzbart, B., and Lynch, H.~J. (2020).
\newblock Sealnet: A fully-automated pack-ice seal detection pipeline for
  sub-meter satellite imagery.
\newblock {\em Remote Sensing of Environment}, 239:111617.

\bibitem[Gotovos et~al., 2013]{Gotovos2013}
Gotovos, A., Casati, N., Hitz, G., and Krause, A. (2013).
\newblock Active learning for level set estimation.
\newblock In {\em IJCAI 2013}.

\bibitem[Grant et~al., 2020]{Grant2020}
Grant, J.~A., Leslie, D.~S., Glazebrook, K., Szechtman, R., and Letchford,
  A.~N. (2020).
\newblock Adaptive policies for perimeter surveillance problems.
\newblock {\em European Journal of Operational Research}, 283(1):265--278.

\bibitem[Guirado et~al., 2019]{Guirado2019}
Guirado, E., Tabik, S., Rivas, M.~L., Alcaraz-Segura, D., and Herrera, F.
  (2019).
\newblock Whale counting in satellite and aerial images with deep learning.
\newblock {\em Scientific reports}, 9(1):1--12.

\bibitem[Heikkinen and Arjas, 1999]{Heikkinen1999}
Heikkinen, J. and Arjas, E. (1999).
\newblock Modeling a poisson forest in variable elevations: a nonparametric
  bayesian approach.
\newblock {\em Biometrics}, 55(3):738--745.

\bibitem[Hsieh et~al., 2018]{Hsieh2018}
Hsieh, Y.-P., Kavis, A., Rolland, P., and Cevher, V. (2018).
\newblock Mirrored langevin dynamics.
\newblock In {\em Advances in Neural Information Processing Systems}, pages
  2878--2887.

\bibitem[John and Hensman, 2018]{John2018}
John, S. and Hensman, J. (2018).
\newblock Large-scale {C}ox process inference using variational {F}ourier
  features.
\newblock In Dy, J. and Krause, A., editors, {\em ICML 2018}, volume~80 of {\em
  Proceedings of Machine Learning Research}, pages 2362--2370,
  Stockholmsmässan, Stockholm Sweden. PMLR.

\bibitem[Kanagawa et~al., 2018]{Kanagawa2018}
Kanagawa, M., Hennig, P., Sejdinovic, D., and Sriperumbudur, B.~K. (2018).
\newblock Gaussian processes and kernel methods: A review on connections and
  equivalences.
\newblock {\em arXiv preprint arXiv:1807.02582}.

\bibitem[Kingman, 1993]{Kingmann2002}
Kingman, J. F.~C. (1993).
\newblock {\em Poisson Processes}.
\newblock Claredon Press.

\bibitem[L\'{o}pez-Lopera et~al., 2018]{Lopez-Lopera2018}
L\'{o}pez-Lopera, A.~F., Bachoc, F., Durrande, N., and Roustant, O. (2018).
\newblock Finite-dimensional gaussian approximation with linear inequality
  constraints.
\newblock {\em SIAM/ASA Journal on Uncertainty Quantification},
  6(3):1224--1255.

\bibitem[L\'{o}pez-Lopera et~al., 2019]{Lopez-Lopera2019}
L\'{o}pez-Lopera, A.~F., John, S., and Durrande, N. (2019).
\newblock Gaussian process modulated cox processes under linear inequality
  constraints.
\newblock {\em arXiv preprint arXiv:1902.10974}.

\bibitem[Maatouk and Bay, 2017]{Maatouk2016}
Maatouk, H. and Bay, X. (2017).
\newblock Gaussian process emulators for computer experiments with inequality
  constraints.
\newblock {\em Mathematical Geosciences}, 49:557--582.

\bibitem[M{\o}ller et~al., 1998]{Moller1998}
M{\o}ller, J., Syversveen, A.~R., and Waagepetersen, R.~P. (1998).
\newblock Log gaussian cox processes.
\newblock {\em Scandinavian journal of statistics}, 25(3):451--482.

\bibitem[Moller and Waagepetersen, 2003]{Moller2003}
Moller, J. and Waagepetersen, R.~P. (2003).
\newblock {\em Statistical inference and simulation for spatial point
  processes.}
\newblock Chapman and Hall/CRC.

\bibitem[Mutn\'{y} and Krause, 2018]{Mutny2018b}
Mutn\'{y}, M. and Krause, A. (2018).
\newblock Efficient high dimensional bayesian optimization with additivity and
  quadrature fourier features.
\newblock In {\em Neural and Information Processing Systems (NeurIPS) 2018}.

\bibitem[Mutn\'{y} and Krause, 2021]{Mutny2021a}
Mutn\'{y}, M. and Krause, A. (2021).
\newblock No-regret algorithms for capturing events in poisson point processes.
\newblock In Meila, M. and Zhang, T., editors, {\em Proceedings of the 38th
  International Conference on Machine Learning (ICML 2021)}, volume 139 of {\em
  Proceedings of Machine Learning Research}, pages 7894--7904. PMLR.

\bibitem[Mutn\'{y} and Krause, 2022]{Mutny2021}
Mutn\'{y}, M. and Krause, A. (2022).
\newblock Sensing cox processes via posterior sampling and positive bases.
\newblock {\em Proceedings of the 25th International Conference on Artificial
  Intelligence and Statistics (AISTATS 2022)}.

\bibitem[Nemirovski and Todd., 2008]{Nemirovski2008}
Nemirovski, A.~S. and Todd., M.~J. (2008).
\newblock Interior-point methods for optimization.
\newblock {\em Acta Numerica}, 17(1):191--234.

\bibitem[Papp and Alizadeh, 2014]{Papp2014}
Papp, D. and Alizadeh, F. (2014).
\newblock Shape-constrained estimation using nonnegative splines.
\newblock {\em Journal of Computational and Graphical Statistics},
  23(1):211--231.

\bibitem[Rasmussen and Williams, 2006]{Rasmussen2006}
Rasmussen, C. and Williams, C. (2006).
\newblock Gaussian processes for machine learning, vol. 1.
\newblock {\em The MIT Press, Cambridge, doi}, 10:S0129065704001899.

\bibitem[Russo, 2016]{Russo2016a}
Russo, D. (2016).
\newblock Simple bayesian algorithms for best arm identification.
\newblock In {\em Conference on Learning Theory}, pages 1417--1418.

\bibitem[Russo et~al., 2017]{Russo2017a}
Russo, D., Van~Roy, B., Kazerouni, A., and Osband, I. (2017).
\newblock A tutorial on thompson sampling.
\newblock {\em arXiv preprint arXiv:1707.02038}.

\bibitem[Shirota et~al., 2017]{Shirota2017}
Shirota, S., Gelfand, A.~E., et~al. (2017).
\newblock Space and circular time log gaussian cox processes with application
  to crime event data.
\newblock {\em The Annals of Applied Statistics}, 11(2):481--503.

\bibitem[Snyder and Miller, 2012]{Snyder2012}
Snyder, D.~L. and Miller, M.~I. (2012).
\newblock {\em Random point processes in time and space}.
\newblock Springer Science \& Business Media.

\bibitem[Srinivas et~al., 2010]{Srinivas2009}
Srinivas, N., Krause, A., Kakade, S.~M., and Seeger, M. (2010).
\newblock Gaussian process optimization in the bandit setting: No regret and
  experimental design.
\newblock {\em International Conference on Machine Learning}.

\bibitem[Vanchinathan et~al., 2015]{Vanchinathan2015}
Vanchinathan, H.~P., Marfurt, A., Robelin, C.-A., Kossmann, D., and Krause, A.
  (2015).
\newblock Discovering valuable items from massive data.
\newblock In {\em Proceedings of the 21th ACM SIGKDD International Conference
  on Knowledge Discovery and Data Mining}, pages 1195--1204.

\bibitem[Virtanen et~al., 2020]{scipy}
Virtanen, P., Gommers, R., Oliphant, T.~E., Haberland, M., Reddy, T.,
  Cournapeau, D., Burovski, E., Peterson, P., Weckesser, W., Bright, J., {van
  der Walt}, S.~J., Brett, M., Wilson, J., Millman, K.~J., Mayorov, N., Nelson,
  A. R.~J., Jones, E., Kern, R., Larson, E., Carey, C.~J., Polat, {\.I}., Feng,
  Y., Moore, E.~W., {VanderPlas}, J., Laxalde, D., Perktold, J., Cimrman, R.,
  Henriksen, I., Quintero, E.~A., Harris, C.~R., Archibald, A.~M., Ribeiro,
  A.~H., Pedregosa, F., {van Mulbregt}, P., and {SciPy 1.0 Contributors}
  (2020).
\newblock {{SciPy} 1.0: Fundamental Algorithms for Scientific Computing in
  Python}.
\newblock {\em Nature Methods}, 17:261--272.

\bibitem[Walder and Bishop, 2017]{Walder2017}
Walder, C.~J. and Bishop, A.~N. (2017).
\newblock Fast bayesian intensity estimation for the permanental process.
\newblock In {\em Proceedings of the 34th International Conference on Machine
  Learning-Volume 70}, pages 3579--3588. JMLR. org.

\bibitem[Williams and Seeger, 2001]{Williams2001}
Williams, C.~K. and Seeger, M. (2001).
\newblock Using the nystr{\"o}m method to speed up kernel machines.
\newblock In {\em Advances in neural information processing systems}, pages
  682--688.

\bibitem[Zhang et~al., 2020]{Zhang2020a}
Zhang, K.~S., Peyré, G., Fadili, J., and Pereyra, M. (2020).
\newblock Wasserstein control of mirror langevin monte carlo.
\newblock {\em Proceedings of The 33rd Conference on Learning Theory, {(COLT
  2020)}}.

\end{thebibliography}
\bibliographystyle{apalike}


\appendix
\newpage
\onecolumn 

 \makesupplementtitle

\etocdepthtag.toc{mtappendix}
\etocsettagdepth{mtchapter}{none}
\etocsettagdepth{mtappendix}{subsection}
\tableofcontents

\section{Extra material}

\subsection{Upper Confidence Bound Algorithm and Laplace Approximation}\label{app:ucb}
A successful class of algorithms for regret minimization are based on the principle of optimism \citep{Auer2002, Srinivas2009}. This principle relies on the access to a confidence set/credibility set, and dictates to sense the region with the largest upper confidence bound (in plots this algorithm has name \textsc{UCB}). 

In order to reason about the confidence bounds without closed form posterior, Laplace approximation offers itself as the first natural candidate. A short calculation reveals that the covariance for the MAP estimate $\hat{\theta}$ with positive basis is $\bSigma_{\text{Laplace},t}  =  \sum_{j=1}^{t}\sum_{i=1}^{n(A_j)} \frac{\Phi(x_i)\Phi(x_i)^\top}{(\Phi(x_i)^\top \hat{\theta} )^2} + \bI$. The upper confidence bound can be then solved as simple convex program (see below). The width of confidence set is inversely proportional to magnitude of $\bSigma_{\text{Laplace}}$. Already before executing this algorithm, we can see that it is destined to fail when $l\approx 0$, since sensing no events $n(A_i) = 0$ does not increase  $\bSigma_{\text{Laplace}}$ at all, and hence does not decrease the confidence region. Thus, we expect the algorithm to get stuck on sensing a region with small counts since it is unable to shrink the confidence estimate for it. 

\citet{Mutny2021} avoids the above described pathology by using a regression estimate instead of maximum likelihood estimate at the cost of blowup to the confidence set by a specifically designed scaling factor that takes into account the tail properties of Poisson distribution. 

\subsection{V-optimal experiment design}\label{app:v-design}
We design an algorithm for approximately learning level sets, and identifying maxima of $\lambda$. We take an inspiration from the literature on statistical experimental design \citep{Chaloner1995,Gotovos2013}, and design an acquisition function based on Bayesian V-optimal design. The general idea is to sequentially reduce the uncertainty in the \emph{region of interest} subject to the cost of sensing.

The \emph{region of interest} $R_t$ depends on our objective. If learning levels sets above a threshold $\tau$, then $R_t = \{ x \in \mD| \operatorname{lcb}_t(x)\geq \tau  \}$, while for identifying maxima, $R_t = \{ x \in \mD | \operatorname{ucb}_t(x) \geq \max_{z\in \mD}\operatorname{lcb}_t(z)\}$. As we proceed with iterations, the region of interest is shrinking to the smallest such set.

The ucb and lcb can be calculated as conic convex optimization problem: 
\begin{equation}\label{eq:ucb}
	\begin{gathered}
		\text{ucb}(A),\text{lcb}(A)  = \max_{\theta} \varphi_A^\top  \theta,\min_{\theta} \varphi_A^\top  \theta \\ ~ \text{subject to} ~  (\theta  - \hat{\theta})^\top \bSigma_{\text{Laplace}}(\theta  - \hat{\theta})\leq \beta(\delta) \\
		 \bGamma \theta \geq l
	\end{gathered}
\end{equation}

\looseness -1 Following ideas of Bayesian experimental design, we greedily select the next sensing region $A \in \mA$ to maximally reduce the posterior squared prediction error in $R_t$  in expectation over the realization of $A$ and in expectation over the posterior over $\theta$ in Equation \eqref{eq:acq1}.

\begin{figure*}[ht]
	\begin{align}
		A_t = \arg\min_{A\in \mA} \frac{1}{w(A)} \mE_{\mathfrak{P}|D}\left[ \int_{\mR^d} \left(\int_{{R}_t}(\Phi(x)(\theta-\hat{\theta}))^2 dx \right) p(\theta|D \cup \{N(A),\{x_i\}_{i=1}^{N(A)} \})d\theta \right]  \label{eq:acq1} \\
		A_t = \arg\min_{A \in \mA} \frac{ \mE_{\mathfrak{P}|\hat{\theta}_A}}{w(A)}\left[ \Tr\left( \int_{R_t} \Phi(x)^\top \Phi(x) dx \left( \sum_{j=1}^t\sum_{i=1}^{N(A_j)} \frac{\Phi(x_i)\Phi(x_i)^\top}{(\Phi(x_i)^\top \hat{\theta}' )^2}  + \bI \right)^{-1}\right) \right]\label{eq:acq2}.
	\end{align}
	\vspace{-0.5cm}
\end{figure*}
	\begin{align}\label{eq:acq}
		{\textstyle	A_t = \argmin_{A} \frac{ \mE_{\mathfrak{P}|\hat{\theta}_A}}{w(A)}\left[ \Tr\left( \int_{R_t} \Phi(x)^\top \Phi(x) dx \left( \bSigma_{\text{Laplace},t-1} + \sum_{i=1}^{N(A_t)} \frac{\Phi(x_i)\Phi(x_i)^\top}{(\Phi(x_i)^\top \hat{\theta} )^2}  \right)^{-1}\right) \right]}
	\end{align}
The expectation in \eqref{eq:acq1} is over the realization of the Poisson process in region $A$ for a constant duration $\Delta$, and $p(\theta|D \cup \{N(A),\{x_i\}_{i=1}^{N(A)} \})$ is the posterior distribution of $\theta$. In general, it is difficult to integrate over the exact posterior, so we again make use of the Laplace approximation of posterior $\theta \sim \mN(\hat{\theta}, \bSigma_{\text{Laplace}})$ to arrive at the simplified acquisition function in Equation \eqref{eq:acq2}. We refer to this algorithm as \textsc{V-optimal}.

The $\hat{\theta}'$ in \eqref{eq:acq2} is a function of $(N(A),\{x_i\}_{i=1}^{N(A)})$ that we take the expectation over, and denotes the new MAP estimate with the new data appended due to sensing $A$. The Poisson process $\mathfrak{P}$ is used to sample $N(A)$ and locations of the points simulating the sensing $A$ with the rate function defined by our current optimistic estimate $\hat{\theta}_A$ for action $A$. The estimator $\hat{\theta}_A$ is defined via $\argmax_\theta \varphi_A^\top \theta$ s.t. it belongs to the same constraint set as in Equation \eqref{eq:ucb}. 

Notice that, as we are interested in identifying maxima or level sets, the optimistic estimate is justified as we hope to discover regions with high values. For example, starting with no information, $\hat{\theta}$ would be close to the lower bound, then sensing under the current MAP estimate $\hat{\theta}$ leads to no event occurrences anywhere except for where it has already seen some samples. This effect is eliminated by instead using an optimistic estimate while simulating sensing in a new region $A$.

In our experiments, we often skip the last steps and set $\theta'=\hat{\theta}$ as the new data does not lead to a significant change in the MAP estimate, leading to \eqref{eq:acq}. The evaluation of the acquisition function in Equation \eqref{eq:acq} requires: calculation of optimistic $\hat{\theta}_A$ (convex program), a sampling step from a Poisson process (sampling), and a MAP calculation with the new sample added (convex program). Sampling from the Poisson process can be done via Monte Carlo simulation as in \citep{Snyder2012}.

\subsection{Derivation of Positive Basis}\label{app:derivations}
In this section we give more information and an alternative argument as to why the basis $\Phi(x) = \bGamma \phi(x)$ needs to be scaled by $\bGamma$ and what is the origin of this scaling. We always start with the basis functions $\phi_i(x)$, which have the desired positive property but are very simple and do not generate the appropriate function class. There are two types of arguments to explain why $\bGamma$ is necessary: a) Bayesian covariance matching (mentioned in the text), and b) Change of basis in Hilbert spaces. 

The Bayesian derivation demands that the prior covariance kernel $k$ is maintained under the basis transformation. In other words $\Phi(x)^\top \Phi(y)^\top \stackrel{!} = k(x,y)$. To achieve this on set of nodes $t_i \in [m]$, we use again $\bV_{ij} = \phi_j(t_i)$, then $\bK = \bV \bGamma \bGamma^\top \bV^\top $. Picking $\bGamma$ to be $\bGamma = \bV^{-1}\bK^{1/2}$ achieves the desired goal. In between the points $t_i$, the values of the kernel $k$ are interpolated.

Nodes $t_i$ needs to be selected such that $\bV$ is invertible (change of basis exists), and this is indeed the case for the bases we consider in our work. In fact, nodes are selected such that $\bV$ becomes close to identity. Similar issues arise also when working Bernstein polynomials and Hermite splines (see \citep{Papp2014}), and not much attention is given to these concepts, since there is always a reasonable choice, however a bad choice of nodes can indeed render the approximation useless. \citet{Papp2014} propose to optimize nodes $t_i$ for Bernstein polynomials but refrain from practically implementing it due to difficulty of the optimization problem. 

For the more Hilbert space inclined reader, the same needs to happen with the RKHS perspective. The optimization in Hilbert spaces proceeds via regularized ERM. The regularizer is the norm of the original RKHS space. If we use a new basis for the RKHS, we need to make sure the same regularization strength $\norm{\lambda}_k^2$ is applied in the new basis. Assuming the domain is finite dimensional only supported on the nodes $t_i$ for $i \in [m]$ we calculate the value of $\norm{\lambda}_k^2$ in the new RKHS. We will see it has form $\theta^\top \bV^{-1} \bK  \bV^{-1} \theta$, where  $\bV_{ij} = \phi_j(t_i)$ is the basis change matrix. The rest follows by using $\Phi(x) = \bV^{-1}\bK^{1/2}\phi(x)$.

\subsection{The square link function: Nodal Lines}\label{app:nodal-lines}
\begin{figure}
	\centering
	\includegraphics[width = 0.7\textwidth]{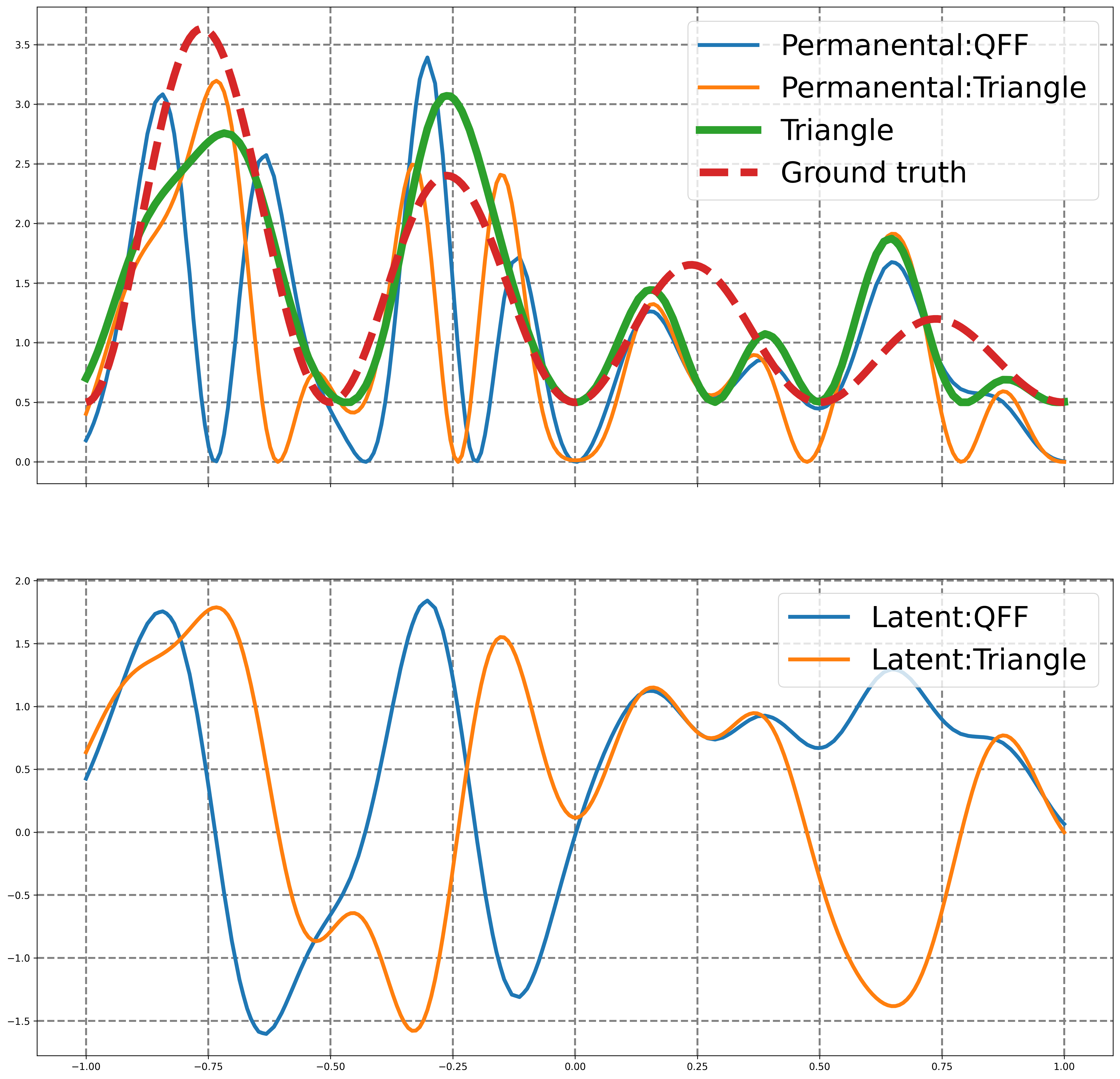}
	\caption{Demonstration of nodal lines with $m = 128$. Green fit is with the constrained formulation and triangle basis.}
	\label{fig:nodal-lines}
\end{figure}

\citet{John2018} pioneered the use of square link function to estimate Poisson process intensity function. The combination of square link function and Gaussian process was later referred to as a Permanental process in literature \citep{Walder2017}. Already \citet{John2018} show that the objective function is non-convex due to equivalence of $-f$ and $f$ where $f$ denotes the latent function passed through the link function. The non-convexity of the estimation problem leads to a phenomenon called \emph{nodal lines}. The nodal lines are direct consequence of non-uniqueness of stationary points in optimization, where a stationary solution to the MAP problem does not fit the solution well in the regions where the latent function $f$ crosses zero. In Figure \ref{fig:nodal-lines}, we plot in the first row estimates of $\lambda$ (in red) with squared link function (blue and orange) as well as our constrained variant (green). We observe that in the second row that once the latent function crosses zero, significant error artefact occurs in estimate of $\lambda$ (in the first plot) which is not present in the case of triangle basis with a constraint. This mainly occurs in the region where $\lambda$ is small but positive as already suggested by \citet{John2018}. Naturally, this effect can be somewhat mitigated by multiple restarts as well as using fixed off-set in the link function, however it turns out to be complicate inference especially when we need to perform multitude of repeated inferences in sequential decision problems. 

%
%
%
\subsection{Other Langevin Dynamics Algorithms}\label{app:langevin}

\subsubsection{Mirrored Langevin Dynamics} \label{app:omd}
Instead of using MYULA - a proximal based Langevin dynamics of \citet{Brosse2017} one could also apply Mirrored Langevin dynamics of \citep{Hsieh2018} due to the simplicity of the constraint.

The Mirrored Langevin dynamics of \citet{Hsieh2018} constructs so called dual distribution $\exp(-W(y))$ which incorporates the constraint in its functional form and is unconstrained such that the standard Langevin dynamics applies. The dual distribution is defined as push-forward via the gradient of the mirror map $\nabla h$. It can be derived by solving Monge-Ampere equations:
\begin{equation}\label{eq:monge-ampere}
 e^{-W} \propto e^{-V(\nabla h^*)} \det (\nabla^2 h^*).
\end{equation}

The mirror map is chosen such that the the dual distribution becomes tractable and convex, and needs to be twice differentiable and strictly convex inside the constraint set. 

The advantage of this algorithm is two-fold it never leaves the constraint set and has better convergence guarantee - it requires only $\mO(\epsilon^2 m)$ to reach $\epsilon$ accuracy under total variation. The disadvantage we found is that its dependence on condition number is worse than MYULA in practice. The condition number becomes proportional to condition number of $\bGamma$, and as we use only gradient based optimization (classical Langevin dynamics) it has bad convergence properties.

We explain the derivation of the mirror map in the steps below. It turns out that our constraint is related to a simple box constraint already analyzed in \citet{Hsieh2018} for which a simple mirror map exists. We reproduce the derivation here for completeness. In order to derive $W$ for our application, we transform the variables to $[-1,1]^m$ domain by applying $\theta = \frac{1}{2}\mathbf{Diag}(\bGamma^{-1} (u-l))z + \frac{1}{2}(\bGamma^{-1}(u+l)) = \bD z + v$ an affine transformation. The potential becomes:

\begin{eqnarray*}
	U(z) = \sum_{j=1}^{t} \sum_{i=1}^{n(A_j)}-\log((\bD z + v)^\top \Phi(x_i)) + (\bD z + v)^\top \varphi_A + \frac{1}{2} \norm{\bD z + v}^2 ~ \text{s.t.} ~ z \in [-1,1]
\end{eqnarray*}

It turns out that the appropriate mirror map for the box constraint is of the form:

\begin{equation}
h(z) = \frac{1}{2}\sum_{i=1}^{m}\left( (1+z_i)\log(1+z_i) +(1-z_i)\log(1-z_i) \right).
\end{equation}

The following mirror map has derivative $\partial h/\partial z_i = \operatorname{arctanh}(z_i)$ and consequently the Fenchel dual has derivative $\partial h^* /\partial y_i = \tanh(y_i)$, where $y_i$ are the dual variables. Using the equation \eqref{eq:monge-ampere} and $\log\det (\nabla^2 h^*) =  \sum_{i=1}^m \log(\frac{1}{2}(1+\cosh(2y_i))$, we can derive the dual distribution up to a constant:
\begin{eqnarray*}
    W(y) \propto \sum_{j=1}^{t} \sum_{i=1}^{n(A_j)}-\log((\bD \tanh(y) + v)^\top \Phi(x_i)) + (\bD \tanh(y) + v)^\top \varphi_A \\+ \frac{1}{2} \norm{\bD \tanh(y) + v}^2 + \sum_{i=1}^m \log(\frac{1}{2}(1+\cosh(2y_i))
\end{eqnarray*}
	
Notice that the $\tanh$ is a monotone function and since composition of convex and monotone remains convex (can be found in \citep{Boyd2004} in Chapter 2), this is a valid objective for Langevin dynamics:
\[ y_{t+1} = y_t - \nabla W(y_t) + \sqrt{2\beta} w_k ~ \text{ where} ~ w_k \sim \mN(0,\bI_m). \]
 The transformation back to primal coordinate is via the dual map $z_T = \nabla h^*(y_T)$. 
 

\subsubsection{Wasserstein Langevin Dynamics}\label{app:wld}
One could also apply a version of Mirrored Langevin Dynamics due to \cite{Zhang2020a}. Akin to the work of \citet{Hsieh2018} it relies on a mirror map to define a constraint but uses different discretization. Let us here describe the problem for a positive constraint with the following classical barrier instead:

\[ h(\theta) = \sum_{i=1}^{m} -\log(\bGamma_{i:}\theta).  \]

The Langevin flow is then defined as follows, 

\[\theta_{k+1} = \nabla h^* \left(\nabla h (\theta_k ) - \eta\nabla U(\theta_k) + \sqrt{\eta \nabla^2 h(\theta)}w_k \right) ~ \text{ where} ~ w_k \sim \mN(0,\bI_m).\]

The inverse of gradient obeys $(\nabla h)^{-1} = \nabla h^*$. With the barrier function above the $(\nabla h)^{-1}$ does not have a closed form and needs to be solved using an iterative process in our case Newton's method. Additionally, using $\eta \propto 1/L$ leads to convergence according to \citet{Zhang2020a}. For further details please refer to the original paper. While this algorithm can avoid bad conditioning; it requires rather costly mirror map inversion procedure. 

\subsubsection{Variational Approximation with P\'olya–Gamma Augmentation}\label{app:donner}
The next possible inference method comes from \citet{Donner2018} which we have learned of during the review process. This method is not general purpose and is specialized for the sigmoid link function. It explicitly uses the form of sigmoid link function in order to represent it as a marginalization over so called P\'olya–Gamma random variables. It then augments the likelihood with these random variable are integrated out. The augmentation also allows for a different interpretation that of marked Poisson process. The joint posterior distribution of augmented variables and the random rate function is approximated with mean field approximation where the two distributions are independent. For further details please refer to the \citep{Donner2018}.

\begin{figure}
	\includegraphics[width=\textwidth]{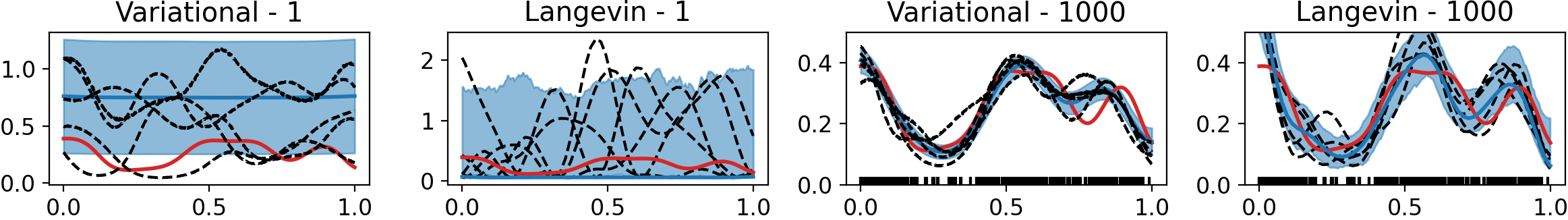}
	\caption{Comparison of variational inference of \citet{Donner2018} with proximal Langevin algorithm of \citet{Brosse2017} primarly used in the experiments for this work. The two plots on the right refer to the prior and the two on the right to the posterior. We see that with the two assumptions priors are different. Also, the approximated inference (variational) leads to somewhat tighter confidence sets on this example due to perhaps the variational approximation. }
	\label{fig:var-polya}
\end{figure}

We depict the different behavior of the variaitonal inference of \citet{Donner2018} and our method in Figure \ref{fig:var-polya}. Apart from having different priors due to the sigmoid link function, we see that the posterior confidence is somewhat tighter, which could be due to approximation or different prior distribution. Also, as formulated by \citet{Donner2018}, the augmentation needs to be done for each sensing region $A_i$ separately which increases the complexity of the method considerably with every new observation. However, as such, can be applied when modeling the sensing problem with sigmoid link function, and generates a decent posterior samples judging from the visual inspection. Similarly to this work, sampling the whole rate function is intractable and the authors resort to sparse GP approximation. 

\subsection{Sigmoid and Hermite basis: why they fail}
The energy function with the log-link function is equal to:

\[ U(\theta) =  \sum_{i=1}^{n(A_j)}\log(u\operatorname{sigmoid}(\theta^\top \Phi(x_i))) -\int_{A_j} u\operatorname{sigmoid}(\theta^\top \bGamma \Phi(x))dx - \frac{1}{2} \norm{\theta}_2^2, \]

where $u$ denotes the maximum value the rate function can attain. Notice that when calculating gradients we need to do approximate quadrature of each dimension of the $\theta_l$:

\[ \partial_l U(\theta) = \sum_{i=1}^{n(A_j)} \frac{\operatorname{sigmoid}'(\theta^\top \Phi(x_i))\Phi_i(x)}{u\operatorname{sigmoid}(\theta^\top \Phi(x_l))}- u\int_{A_j}\operatorname{sigmoid}'(\theta^\top \Phi(x))\Phi_l(x)dx - \theta_l \]

From \citep{Mutny2018b} we know that, while Hermite basis, is optimal in its description size for squared exponential kernel, it has rapidly increasing frequencies $\Phi_l(x) \propto \cos(\omega_l x)$, as $l$ increases in its Fourier spectrum. This means that $\Phi_l(x)$ is a oscillatory function with high frequency, which are notoriously difficult to integrate \citep{Davis2007}. Hence, the approximate inference scheme which we employ: Legendre quadrature with fixed nodes, fails. Naturally, an adaptive quadrature scheme for each frequency could improve the convergence of this method, however already with our fixed quadrature rule it is significantly more costly, and not competitive in terms of computational time. This does not happen with triangle basis since each basis function $\Phi_l(x)$ is a simple piece-wise linear function, hence the performance is unhampered with Legendre quadrature. 

\textbf{This demonstrates a very important point. While the basis representation $\Phi$, has no effect on the function \emph{values}, it does have an important effect on \emph{inference scheme}. Namely, in this case it influences the regularity conditions of the gradient of the loss function with respect to the parametrization. }

\section{Numerical Experiments}\label{app:experiments}
We run our experiments on a local 28 core machine with 128GB memory for ca. 5 days total runtime. Individual operations are not extremely costly but need to repeated many times due to sequential nature of the algorithms. This significantly blows up the computational time.

\paragraph{Algorithmic Choices}

\begin{enumerate}
	\item We perform sampling with the proximal algorithm of \citet{Brosse2017} for regret minimization with $1000$ steps, and Lipschitz constant calculated at each step approximately. We always initialize with the MAP estimate. The number of iterations of the iterative procedure is set such that the samples visually appear to capture the posterior variation of the rate function adequately. All our examples are either one or two dimensional, and samples from the posterior can be easily visually inspected to determine if the approximate sampling produces viable samples. Naturally increasing this number may improve the validity of the samples, but seems to be not required for our application. Also, notice an approach with link function requires the same choice to be made. 	
	\item We create $|\mA|$ - set of sensing regions - by hierarchical splitting the domain with the size for specific problem given below. 
	\item Both \textsc{UCB} algorithms and \textsc{V-optimal}, we use $\beta = 3$. \textsc{V-optimal} resamples from the Poisson process $10$ times to calculate the acquisition function. We use the algorithm of \citet{Snyder2012} to sample on a discretized domain. 
	\item For approximate inference we use Legendre quadrature of scipy \citep{scipy}
	\item In the table below one can find specifics for each dataset. 
\end{enumerate}
\subsection{Fit and test approach}
We want to evaluate the applicability of our approach without the model misspecification interfering in understanding of the performance, and at the same time be able to include randomness of the process into our algorithmic comparison. We execute the experiments in \emph{fit and test} approach. We first fit the model to the whole data corresponding to our application that results in a ground truth rate function $\lambda$. In all cases, this is visually satisfactory fit as can be seen in the Figures below. We then perform the experiments, where points are sampled from the fitted $\lambda$ and compared with respect to it as the ground truth. This way we will avoid overfitting to a specific data realization in the data, which can lead to missrepresentation of performance due to specifics of the instance. 

\subsection{Parameters}
In the table below SE denotes squared exponential kernel.
\begin{center}
	\begin{tabular}{|p{2cm}|c|p{1.2cm}|p{2cm}|p{1cm}|c|c|c|c|c|}\hline
		\textbf{Experiment name} & \textbf{Domain} & \textbf{Kernel} & \textbf{Hyper-parameters} & $l$ & $m$&$\tau$ &  $\Delta$ & $|\mA|$ & T  \\ \hline
		\emph{toy-problem} & $[-1,1]$ & SE & lengthscale $\gamma = 0.1$ & $0.1$ & $64$& &  $5$ & 128 & 400\\
		\emph{San Francisco dataset} & $[-1,1]^2$ & custom & lengthscale $\gamma = 0.1$, (see below) & $0.05$  & $36$& & $30$ days & 64 & 200\\ 
		\emph{Cholera dataset} & $[-1,1]^2$ & custom &lengthscale $\gamma = 0.5$ (see below) & $0$ & $100$& & $ 1/20$ of d.d. & 256 & 50\\
		\emph{Beilschmiedia dataset} & $[-1,1]^2$ & custom &lengthscale $\gamma = 0.1$ (see below) & $0.01$  & $100$& $525/2$ & $ 1$ & 64 & 100\\
		\emph{Antartic seal dataset} & $[-1,1]^2$ & custom &lengthscale $\gamma = 0.25$ (see below) & $0.01$ &  $20^2$ &$350$& $1$ & 64 & 60\\ \hline
	\end{tabular}
\end{center}
By d.d, we mean dataset duration which means the time span between first and last event. By the ratio of dataset we mean that if the time duration of the dataset was lets say $30$ days and the ratio was $1/2$, this means that one sensing session was  $\Delta = 15$ days. Further details are provided below.

\paragraph{Caution: Simplifications!}
Before we describe the benchmarks, we stress that while the experiments stem from real problems, they are simplified versions. Each application dictates a lot of additional technological, ethical and economical specifications that we do not consider. Therefore, one should see the following as show-cases of what is possible, rather than the complete solutions. More crucially, some of the benchmarks can be improved by expert designed kernels $k$. For example, ecologists might have prior knowledge regarding the seals' habitat based on geographical features, and this should be incorporated to $k$ improve efficiency. With the lack of such expert knowledge, we rely only on naive geographical features and spatial correlation. However, the strength of this formalism lies in its flexibility. 

\subsubsection{The San Francisco benchmark} The data comes for this experiment comes from  \url{https://data.sfgov.org/Public-Safety/Police-Department-Incident-Reports-Historical-2003/tmnf-yvry}. The dataset contains much more information that we do not use. We just use data for $880$ days of bulglary events in San Francisco. We model the problem using a kernel 
$ k(x,y) = \exp(-\norm{x-y}_2^2/\gamma^2) w(x)w(y)$,
where $w(x)$ indicates whether the given coordinate is urban on not with $0$ if not. This way we disregard the portions of the map which correspond to the forests or water areas. The lenghtscale was determined by visually inspecting the data and picking an appropriate lenghtscale such that the fit corresponded to meaningful rate. Alternative approaches include is to integrate the uncertainty over the posterior and maximize the Bayesian evidence, which in this case is difficult due to absence of closed form of the posterior. Note that the image in the banner includes a cut-off on the rate function to avoid clutter.

\begin{center}
	\includegraphics[width=0.6\textwidth]{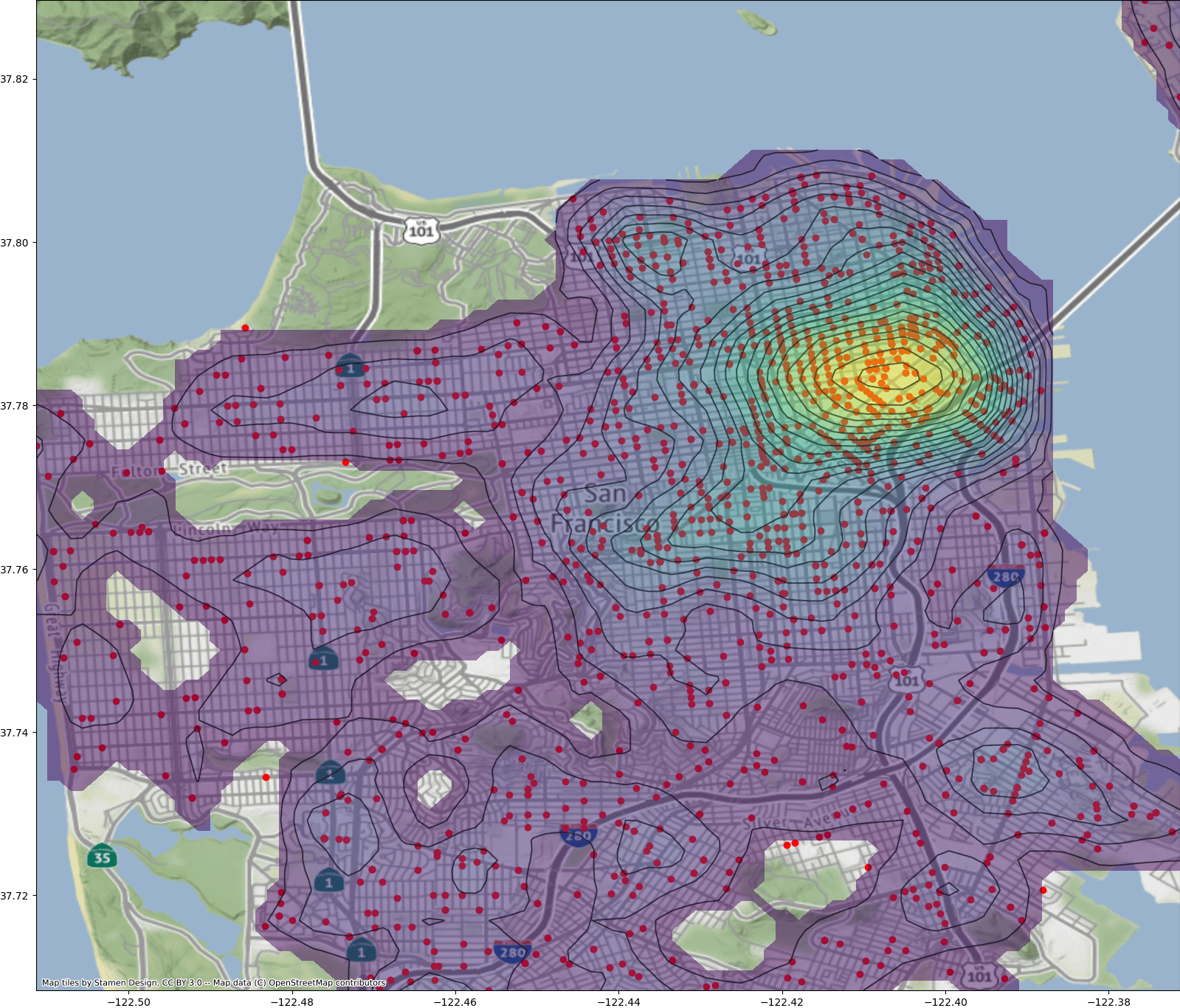}
\end{center}

\subsubsection{Cholera benchmark} 
The next dataset is a classic example of pollutant modeling, and relies on historical dataset. During the London cholera outbreak of 1854, John Snow used records of cholera deaths to identify a contaminated water well -- which is nowadays regarded as the cradle of spatial epidemiology. Assuming that contamination propagates locally from the source, infections can be modeled as Poisson point process as in Fig. below with water wells (blue) as inducing points. The goal in this example is to stop an epidemic as soon as possible and identify the contaminant by surveying the neighbourhoods -- sensing -- to identify sick individuals as the epidemic evolves. The problem is modelled with SE (squared exponential kernel) with inducing point in the water wells according to the strategy in \citet{Williams2001}.

\begin{center}
\includegraphics[width=0.6\textwidth]{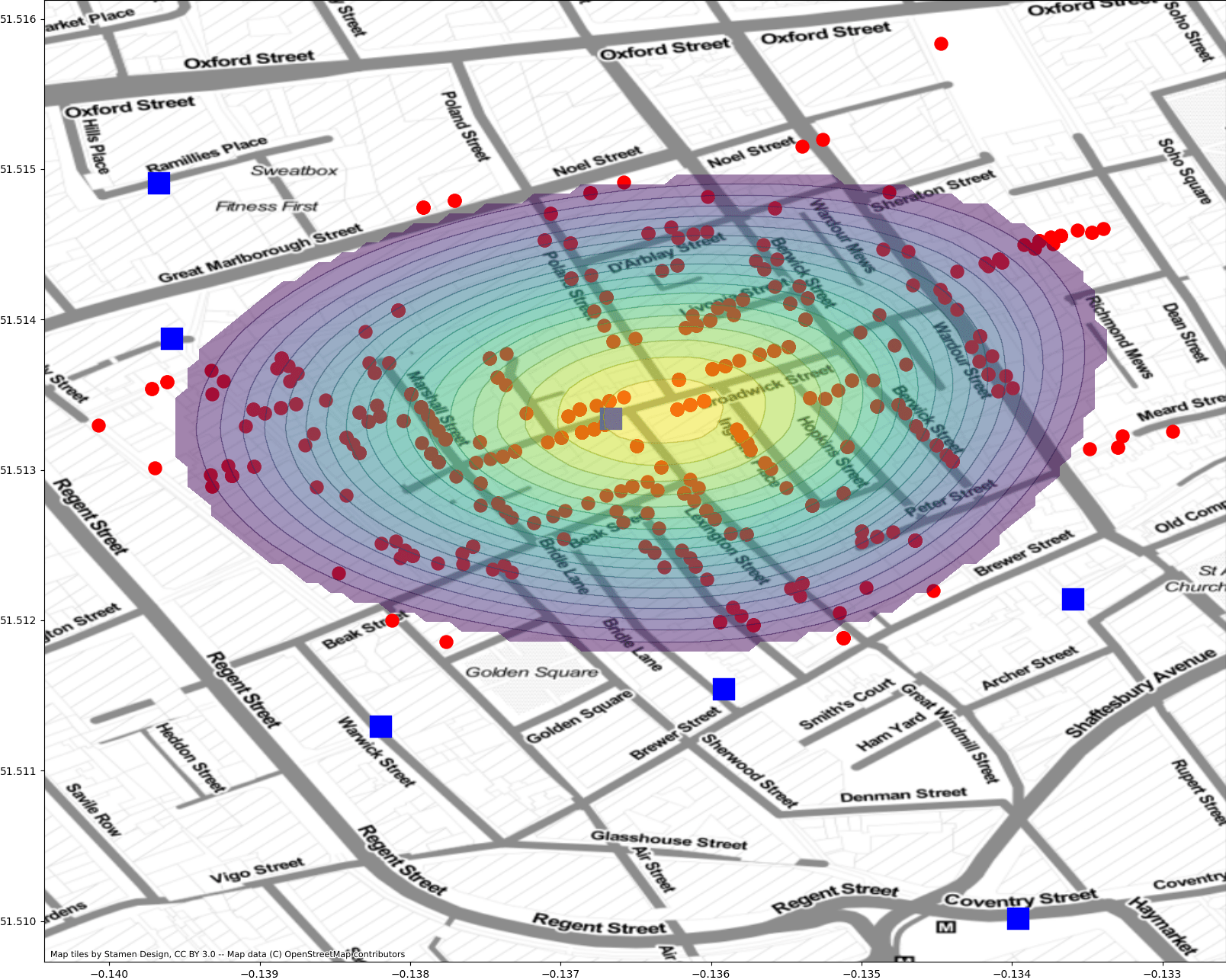}
\label{fig:cholera}
\end{center}

\subsubsection{Beilschmiedia benchmark} This dataset comes from the work of \citet{Baddeley2015}, and we run level set identification $\tau = 525/2$, which covers approximately $20\%$ of the domain. In modelling we use kernel $k((x_1,y_1),(x_2,y_2)) = \exp( -(s_{x_1,y_1} - s_{x_2,y_2})/\gamma^2)\exp( -(h_{x_1,y_1} - h_{x_2,y_2})/\gamma^2)$, where $s_{x,y}$ corresponds to the slope magnitude at $x,y$ and $h_{x,y}$ to the height at $x,y$ for which we have look-up tables. While this representation is not perfect for the observed data it matches them quite closely for sufficiently good fit. In this example, we did not do any elaborate hyperparameter fitting.

\begin{center}
	\includegraphics[width=0.6\textwidth]{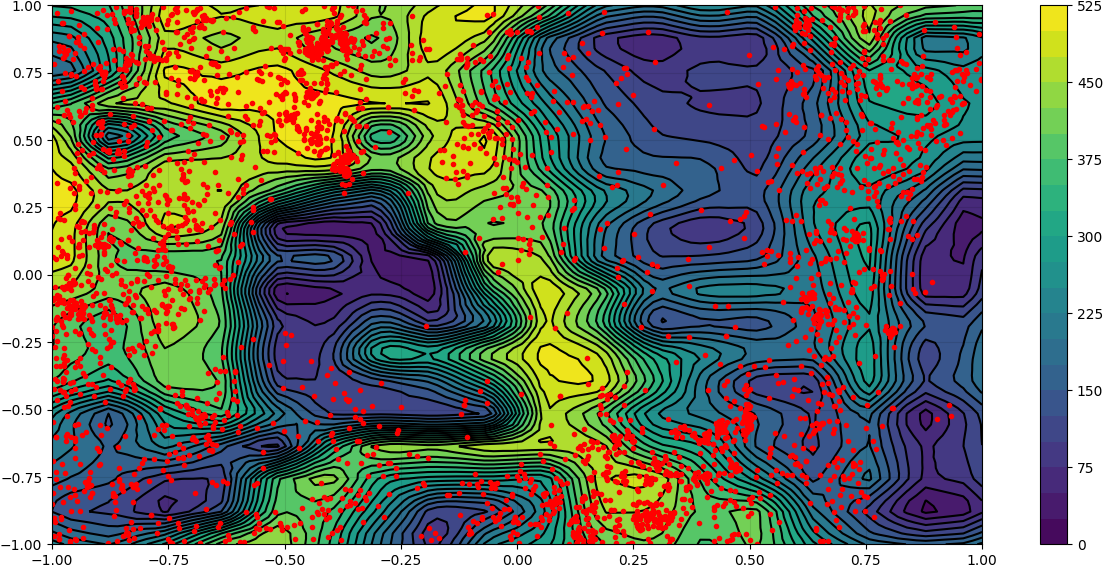}
\end{center}

\subsubsection{Antartic seal benchmark} This dataset comes from \citet{Goncalves2020}. The dataset contains images and locations of identified seals. We do not use the images in this work. We use only locations in the training set to fit a Cox process on a section of antarctic around Ross sea. We use kernel $k(x,y) = \exp(-\norm{x-y}_2^2/\gamma^2)w(x)w(y)$ where $w(x)$ indicates whether the point x is at most $200$km away from the Antarctic continent. Notice from the image that Artactic continent and Antartic coast are not the same. This modelling is purely arbitrary and could significantly benefit from expert knowledge. However, already such simplified setting can provide interesting insights to efficiency of the tested adaptive algorithms. We use as $\tau = 300$ which is approximately half of the maximum rate function. 

\begin{center}
	\includegraphics[width=0.6\textwidth]{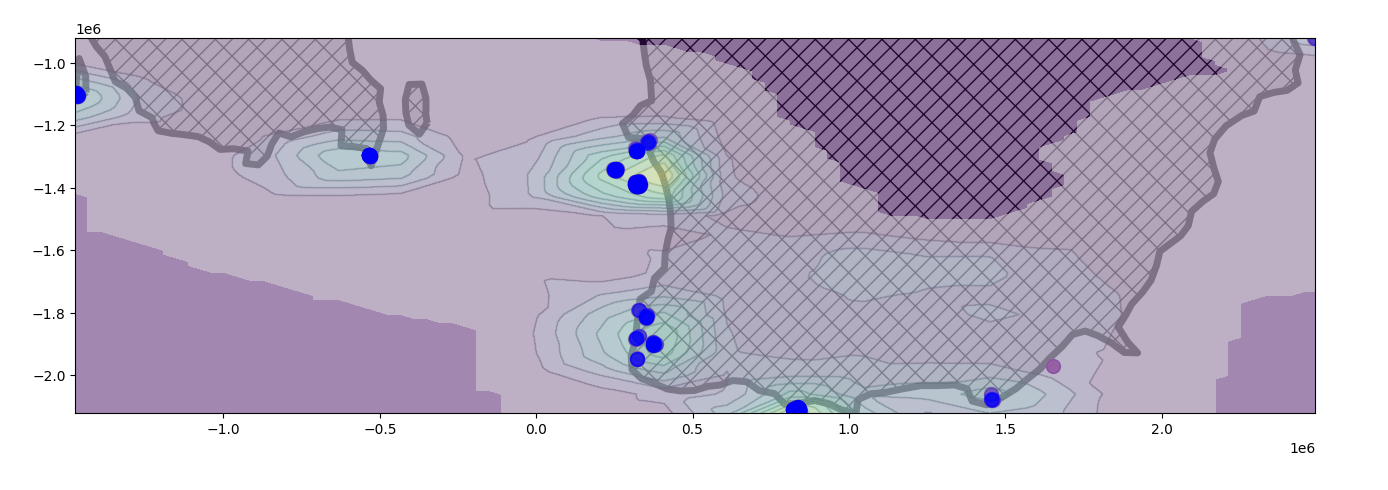}
\end{center}

\subsection{Further experiments: concave cost and runtime}
\begin{figure}[h]
	\centering
	\begin{subfigure}[t]{0.45\textwidth}
		\includegraphics[width = \textwidth]{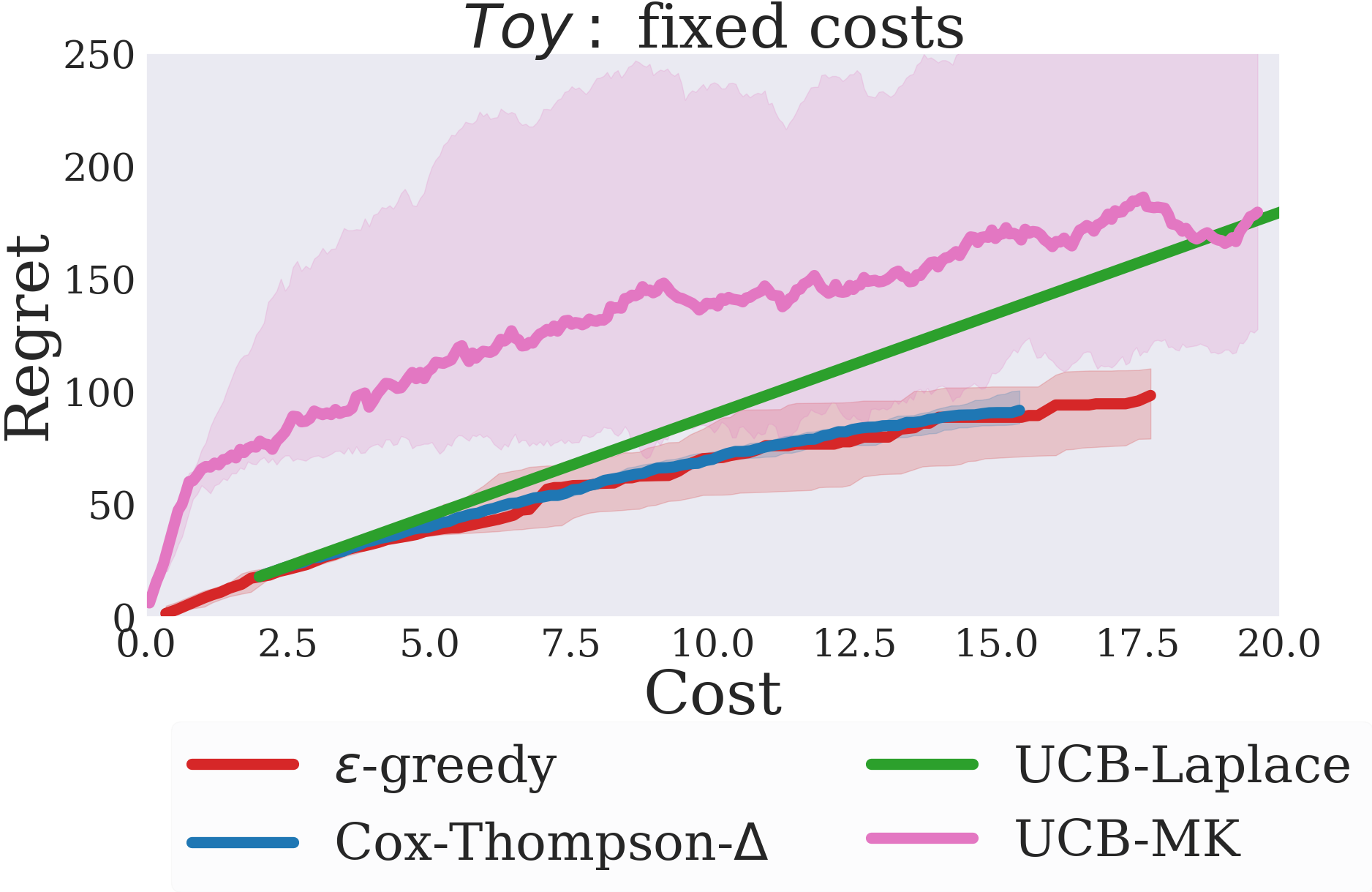}
		\caption{\emph{Toy problem}: fixed costs $w(A) = |A| + 0.02$. Notice that some algorithms do not start at zero cost since they query a large set at first to make use of the budget efficiently. Also, note that the ordering of UCB and Thompson sampling is reversed.}
		\label{fig:1d_count-record_concave}
	\end{subfigure}
~
	\begin{subfigure}[t]{0.45\textwidth}
	\includegraphics[width = \textwidth]{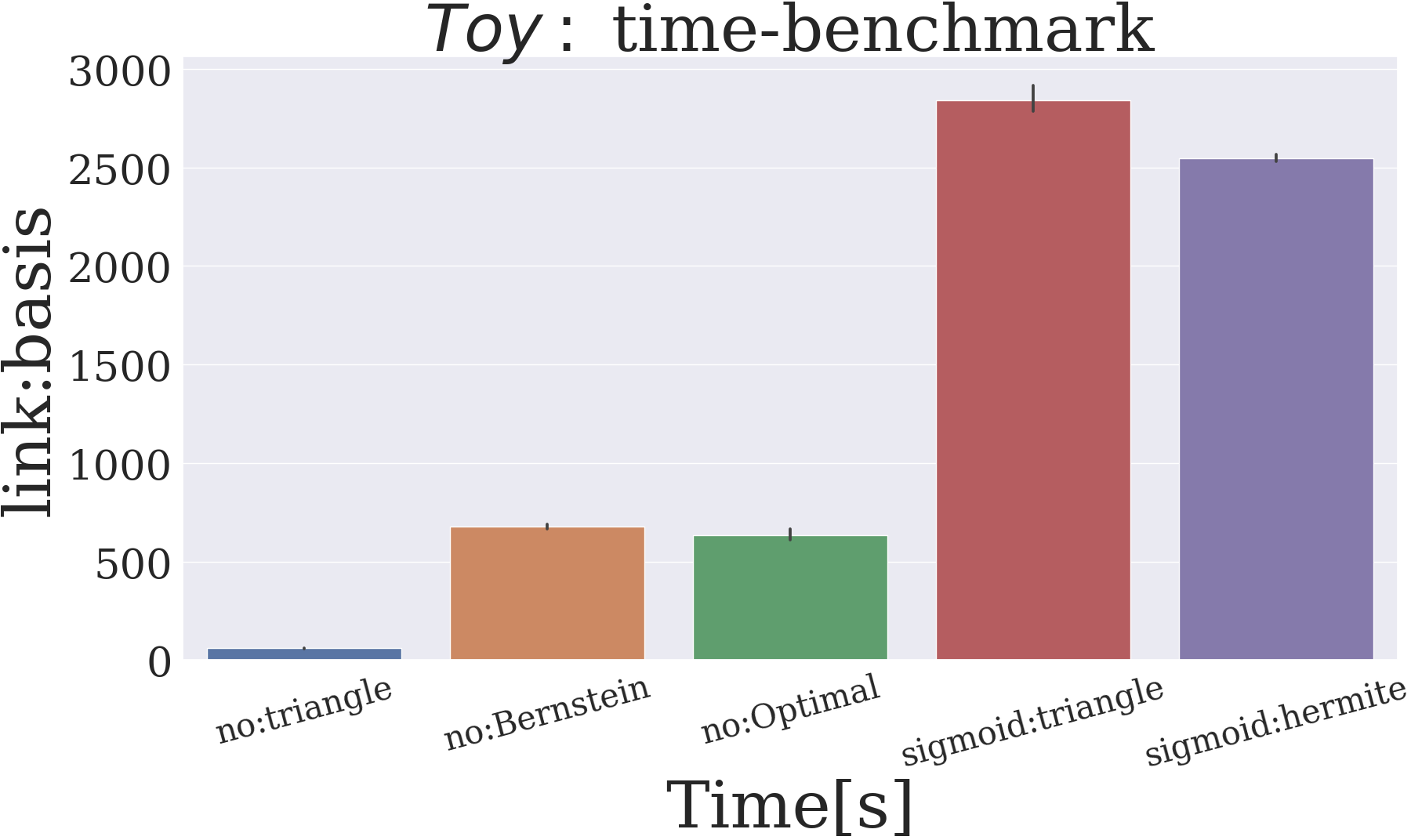}
	\caption{\emph{Toy problem}: Time analysis of experiment in \ref{fig:experiments} a). Standard deviation reported in error bars. We see that link function is significantly more costly in terms of computational time.}
	\label{fig:1d_time}
\end{subfigure}
	\caption{Further experiment with histogram feedback and computational time analysis.}
	\label{fig:extra}
\end{figure}

Not all our experiments could fit to the main body of the paper. We report two more experiment in Figure \ref{fig:extra} with \emph{fixed costs} (simplest version of concave cost), as well as time benchmark for the different link functions. 

\subsection{How to pick $m$?}
The basis approximation we discuss in our work requires the specification of parameter $m$ -- basis size. With increasing basis size, the approximation has higher fidelity, however the computational cost grows as $\mO(m^2)$ (matrix-vector multiplication). Hence, we want smallest basis which still captures the complexity of the problem. One way to determine if $m$ is sufficiently big is to do it by visual inspection, i.e. looking at the samples of the Bayesian model with fixed $m$ and seeing if the samples have sufficient variation we believe the studied problem should have. 

Secondly, given a certain cut-off $\epsilon$ on the eigenvalue spectrum, we can calculate the eigenvalues of the Gram matrix $\bK$ on the selected nodes $t_i$. We can adopt the following procedure: Consider $t_i$ which are on a regular grind in $d$ (one or two) dimensions, which are generated by Cartesian products. We gradually increase $m$ and hence the granularity of the grid. For each value of $m$, we calculate the eigenvalues of $\bK$, namely the minimum eigenvalue. If the minimum eigenvalue is below the cut-off value, we can declare the the approximation to be of sufficient fidelity. The exact relationship between e.g. $L_\infty$ and cut-off $\epsilon$ is not straightforward but follows an inverse relationship, and can be empirically established. 

\subsection{Influence of $\beta$ for UCB}
\begin{figure}[h]
	\centering
	\includegraphics[width=0.5\textwidth]{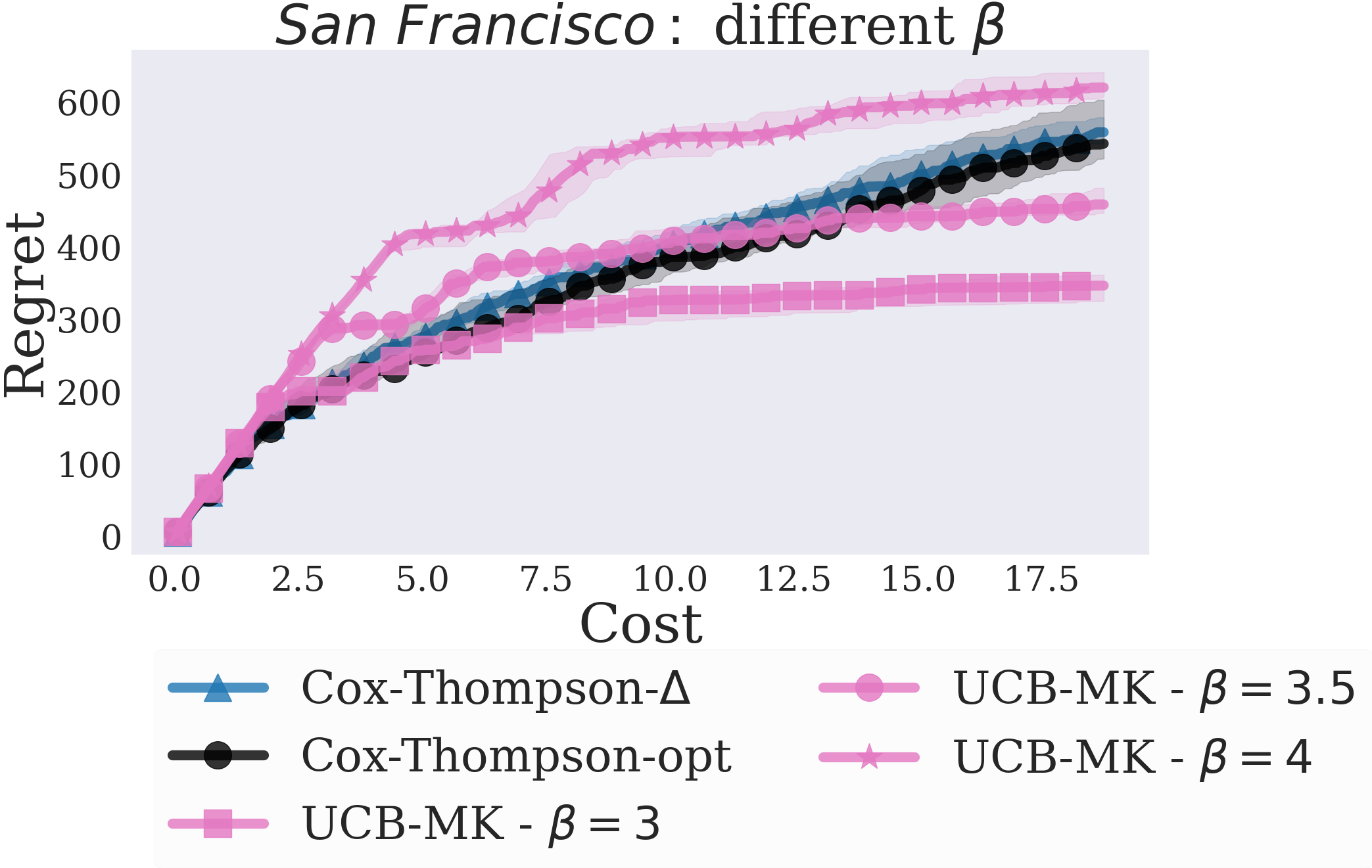}
	\caption{San Francisco experiment: Comparison of UCB-MK with different values of $\beta$.}
	\label{fig:beta}
\end{figure}

In Fig.~\ref{fig:experiments}c) it seems like UCB of \citet{Mutny2021} outperforms Thompson sampling. While it is possible that on this specific problem UCB performs better, it is more the cause of optimized value of confidence parameter $\beta$. In Fig.~\ref{fig:beta} we report the same experiment with multiple values of $\beta$ and we see that the performance significantly deteriorates with different choice of $\beta$. Also, note that the values of $\beta$ as specified by the theory are much larger than used in the comparison here. Naturally, in this problem it seems that low beta due to unimodality of the objective is optimal, since by detecting the first peak the best strategy is to focus on this peak solely and stop exploring. However, this is a special property of this benchmark only. 

The value of $m = 64$ in this experiment.

\end{document}